\pdfoutput=1

\documentclass[11pt]{article}

\usepackage[hyperref]{acl}

\usepackage{times}
\usepackage{latexsym}

\usepackage{graphicx}
\graphicspath{ {./images/} }

\usepackage[T1]{fontenc}

\usepackage[utf8]{inputenc}

\usepackage{microtype}

\usepackage{booktabs}
\usepackage{xcolor}
\usepackage{amsmath}
\usepackage[frozencache,cachedir=.]{minted}

\usepackage{enumitem} 
\setlist[itemize]{noitemsep,topsep=0pt} 

\definecolor{SGcolor}{rgb}{0.3,0.9,0.3}
\definecolor{JNcolor}{rgb}{0.8,0.1,0.1}

\title{GEMv2: Multilingual NLG Benchmarking in a Single Line of Code}

\author{
\begin{minipage}[t]{\textwidth}
\centering
\small
Sebastian Gehrmann$^{11}$,
Abhik~Bhattacharjee$^{3}$,
Abinaya~Mahendiran$^{24}$,
Alex~Wang$^{25}$,
Alexandros~Papangelis$^{2}$,
Aman~Madaan$^{4}$,
Angelina~McMillan-Major$^{15}$,
Anna~Shvets$^{10}$,
Ashish~Upadhyay$^{32}$,
Bingsheng~Yao$^{31}$,
Bryan~Wilie$^{39}$,
Chandra~Bhagavatula$^{1}$,
Chaobin~You$^{41}$,
Craig~Thomson$^{43}$,
Cristina~Garbacea$^{47}$,
Dakuo~Wang$^{20,26}$,
Daniel~Deutsch$^{48}$,
Deyi~Xiong$^{41}$,
Di~Jin$^{2}$,
Dimitra~Gkatzia$^{8}$,
Dragomir~Radev$^{51}$,
Elizabeth~Clark$^{11}$,
Esin~Durmus$^{34}$,
Faisal~Ladhak$^{7}$,
Filip~Ginter$^{49}$,
Genta~Indra~Winata$^{39}$,
Hendrik~Strobelt$^{16,20}$,
Hiroaki~Hayashi$^{4,33}$,
Jekaterina~Novikova$^{50}$,
Jenna~Kanerva$^{49}$,
Jenny~Chim$^{29}$,
Jiawei~Zhou$^{14}$,
Jordan~Clive$^{6}$,
Joshua~Maynez$^{11}$,
João~Sedoc$^{25}$,
Juraj~Juraska$^{44}$,
Kaustubh~Dhole$^{9}$,
Khyathi~Raghavi~Chandu$^{22}$,
Laura~Perez-Beltrachini$^{45}$,
Leonardo~F.~R.~Ribeiro$^{38}$,
Lewis~Tunstall$^{15}$,
Li~Zhang$^{48}$,
Mahima~Pushkarna$^{11}$,
Mathias~Creutz$^{46}$,
Michael~White$^{40}$,
Mihir~Sanjay~Kale$^{11}$,
Moussa~Kamal~Eddine$^{53}$,
Nico~Daheim$^{30}$,
Nishant~Subramani$^{1,21}$,
Ondrej~Dusek$^{5}$,
Paul~Pu~Liang$^{4}$,
Pawan~Sasanka~Ammanamanchi$^{17}$,
Qi~Zhu$^{42}$,
Ratish~Puduppully$^{45}$,
Reno~Kriz$^{18}$,
Rifat~Shahriyar$^{3}$,
Ronald~Cardenas$^{45}$,
Saad~Mahamood$^{52}$,
Salomey~Osei$^{21}$,
Samuel~Cahyawijaya$^{13}$,
Sanja~Štajner$^{35}$,
Sebastien~Montella$^{27}$,
Shailza~Jolly$^{37}$,
Simon~Mille$^{28}$,
Tahmid~Hasan$^{3}$,
Tianhao~Shen$^{41}$,
Tosin~Adewumi$^{19}$,
Vikas~Raunak$^{23}$,
Vipul~Raheja$^{12}$,
Vitaly~Nikolaev$^{11}$,
Vivian~Tsai$^{11}$,
Yacine~Jernite$^{15}$,
Ying~Xu$^{47}$,
Yisi~Sang$^{36}$,
Yixin~Liu$^{51}$,
Yufang~Hou$^{16}$
 \\
{\footnotesize \normalfont
$^{1}$Allen Institute for AI,
$^{2}$Amazon Alexa AI,
$^{3}$Bangladesh University of Engineering and Technology,
$^{4}$Carnegie Mellon University,
$^{5}$Charles University,
$^{6}$Chattermill,
$^{7}$Columbia University,
$^{8}$Edinburgh Napier University,
$^{9}$Emory University,
$^{10}$Fablab in Paris by Inetum,
$^{11}$Google Research,
$^{12}$Grammarly,
$^{13}$HKUST,
$^{14}$Harvard University,
$^{15}$Hugging Face,
$^{16}$IBM Research,
$^{17}$IIIT Hyderabad,
$^{18}$Johns Hopkins University,
$^{19}$Luleå University of Technology,
$^{20}$MIT-IBM Watson AI Lab,
$^{21}$Masakhane,
$^{22}$Meta AI,
$^{23}$Microsoft,
$^{24}$Mphasis NEXT Labs,
$^{25}$New York University,
$^{26}$Northeastern University,
$^{27}$Orange Labs,
$^{28}$Pompeu Fabra University,
$^{29}$Queen Mary University of London,
$^{30}$RWTH Aachen University,
$^{31}$Rensselaer Polytechnic Institute,
$^{32}$Robert Gordon University,
$^{33}$Salesforce Research,
$^{34}$Stanford University,
$^{35}$Symanto Research,
$^{36}$Syracuse University,
$^{37}$TU Kaiserslautern,
$^{38}$Technical University of Darmstadt,
$^{39}$The Hong Kong University of Science and Technology,
$^{40}$The Ohio State University,
$^{41}$Tianjin University,
$^{42}$Tsinghua University,
$^{43}$University of Aberdeen,
$^{44}$University of California, Santa Cruz,
$^{45}$University of Edinburgh,
$^{46}$University of Helsinki,
$^{47}$University of Michigan,
$^{48}$University of Pennsylvania,
$^{49}$University of Turku,
$^{50}$Winterlight Labs,
$^{51}$Yale University,
$^{52}$trivago N.V.,
$^{53}$École Polytechnique\\
}
{\small \normalfont \texttt{gehrmann@google.com, gem-benchmark@googlegroups.com}}
\end{minipage}
}

\begin{document}
\maketitle
\begin{abstract}
Evaluation in machine learning is usually informed by past choices, for example which datasets or metrics to use. This standardization enables the comparison on equal footing using leaderboards, but the evaluation choices become sub-optimal as better alternatives arise. This problem is especially pertinent in natural language generation which requires ever-improving suites of datasets, metrics, and human evaluation to make definitive claims.
To make following best model evaluation practices easier, we introduce GEMv2. The new version of the Generation, Evaluation, and Metrics Benchmark introduces a modular infrastructure for dataset, model, and metric developers to benefit from each others work. GEMv2 supports 40 documented datasets in 51 languages. Models for all datasets can be evaluated online and our interactive data card creation and rendering tools make it easier to add new datasets to the living benchmark.
\end{abstract}

\begin{figure*}[ht]
\centering
\includegraphics[width=0.65\textwidth]{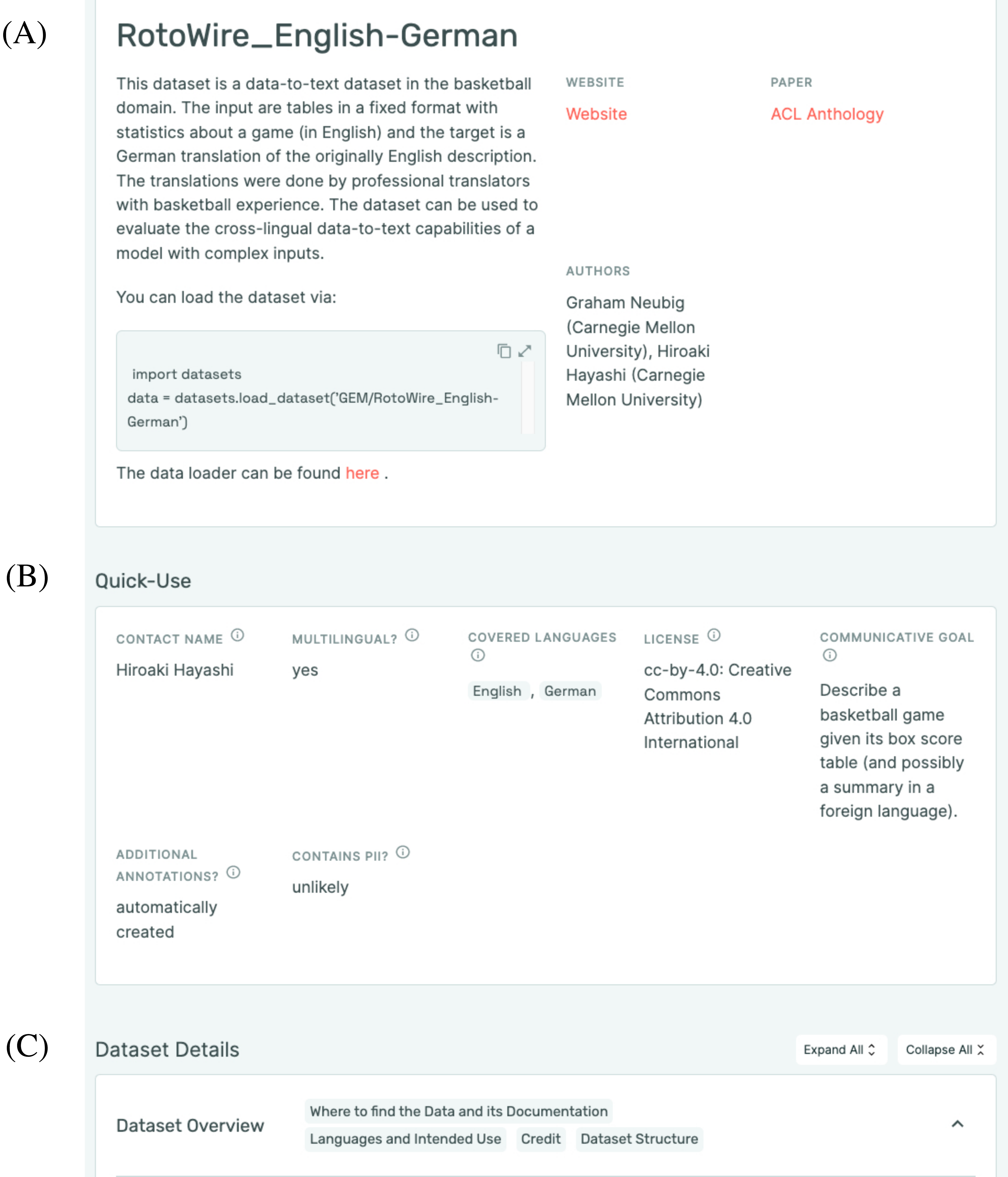}
\caption{One of the data cards for GEM datasets. (A) shows the header which has the name, a summary, a code example, and links to the loader and relevant papers and websites, alongside an author list. (B) is the Quick-Use section which summarizes the most important aspect of a dataset, including language(s), PII, and licensing information. (C) is the detailed view which has multiple sections like ``Dataset Overview''. Each section provides a glance at categories of included questions, and expands to full details on click.}
\label{fig:datacard}
\end{figure*}

\section{Introduction}

The standard evaluation process in natural language processing involves comparisons to prior results in a fixed environment, often facilitated through benchmarks and leaderboards.
This process, if executed correctly, can advance reproducibility~\citep{belz-etal-2021-systematic} and standardize evaluation choices that lead to better dataset diversity. But static benchmarks also prevent the adoption of new datasets or metrics~\citep{DBLP:conf/nips/RajiDBHP21}, and many evaluation advancements are thus put aside.
That means that the focus on surpassing the best prior reported scores reinforces outdated evaluation designs. Furthermore, this process ignores properties that do not match the leaderboard metric~\citep{ethayarajh-jurafsky-2020-utility,bowman-dahl-2021-will,DBLP:journals/corr/abs-2107-07002}.
This issue is particularly pertinent in natural language generation (NLG) since the model quality cannot be estimated using accuracy and instead, NLG relies on automatic and human evaluation approaches that constantly improve~\citep{gehrmann2022repairing, Kasai2021BidimensionalLG}.

To bridge the gap between advantages of leaderboards and in-depth and evolving evaluations, the Generation, Evaluation, and Metrics benchmark~\citep[GEM,][]{gehrmann-etal-2021-gem} proposed a ``living'' benchmark.
As such, GEM is participatory in that contributors propose new datasets and expand the selection of metrics. Model developers using GEM retain full agency over the evaluation process but are able to choose from a wider range of tasks and metrics. GEM further introduced evaluation suites~\citep{mille2021automatic, dhole2021nl} that are compatible with its datasets and test various robustness and fairness aspects of models.

We uncovered several shortcomings in GEMv1 that hindered its scaling and adoption: (1) Centralized data management made adding new datasets too complex. (2) Computing all metrics in a single framework led to dependency issues and was challenging for those with limited compute resources. (3) Participants needed more guidance in our dataset documentation process~\citep{mcmillan-major-etal-2021-reusable} to guarantee data card quality.

We introduce GEMv2, a modular and extendable NLG evaluation infrastructure which allows for continuous integration of newly developed datasets. We release a data card collection and rendering tool that makes it easier to follow for both card creators and readers.
These improvements led to an expansion of GEM from 13 to 40 tasks and from 18 to 51 supported languages.
We also introduce an online evaluation process that collects model outputs and computes metrics for all datasets.

\section{Features and Functionality}

Since best evaluation practices change over time, the infrastructure is modular and maintainable and allows for dataset and metrics additions so they are compatible with all other features. Model developers are able to use new datasets and metrics without any changes to their existing setup. In this section, we describe the supported user [\textit{J}]ourneys for various stakeholders in generation research.

\noindent \textbf{J1 - Document a Dataset} Documentation is a requirement for any dataset in GEM. Our data card template is based on that by \citet{mcmillan-major-etal-2021-reusable}, which was revised using the Data Card Playbook~\citep{pushkarna2021datacardsplaybookupdated}. A data card can be filled out via an interactive form that provides instructions for each field to account for differences in expertise of the documentation writers.\footnote{\url{huggingface.co/spaces/GEM/DatasetCardForm}} The form can load existing data cards to make updates.

\noindent \textbf{J2 - Choose a Dataset} The data card viewer presents information at multiple levels of details in separate columns. Anyone can quickly get a high-level overview of a dataset to make an appropriate selection, or look for detailed information on a documentation category (see Figure~\ref{fig:datacard}).

\noindent \textbf{J3 - Create a Data Loader} Each dataset has a separate repository at \url{huggingface.co/GEM}, with a loader using the Datasets library~\citep{lhoest-etal-2021-datasets}.\footnote{Documentation on how to add new datasets can be found at \url{gem-benchmark.com/tutorials}.} Through this, all supported datasets can be loaded via the same code,

\begin{minted}[fontsize=\footnotesize,autogobble,frame=single]{python}
  from datasets import load_dataset
  data = load_dataset(
      'GEM/$dataset_name',
      '$config_name')
\end{minted}

\noindent where \texttt{\$dataset\_name} is the name of the dataset and \texttt{\$config\_name} is the (optional) specification of the dataset configuration to use.
To stratify how datasets are accessed, they are implemented according to the following conventions:
\begin{itemize}
    \item \texttt{linearized\_input}: Linearization processes convert structured input to a string. For reproducibility, we implement linearization schemes following earlier work \citep[e.g.,][]{saleh-etal-2019-naver,kale-rastogi-2020-text,puduppully-lapata-2021-data}.
    \item \texttt{target} and \texttt{references}: To make all datasets compatible with standard training and evaluation schemes, all datasets have a string target and a list of string references field.
    \item \texttt{gem\_id}: To be able to track outputs even for shuffled datasets, each GEM dataset assigns a unique ID to all examples, which the evaluation library uses to unshuffle.
\end{itemize}

\noindent \textbf{J4 - Evaluate a Model} 
Model outputs can be evaluated locally using the \texttt{gem-metrics} library or online which will add the outputs to our result overview (J6).\footnote{\url{huggingface.co/spaces/GEM/submission-form}}
Both methods require a standardized input format that specifies the dataset and split and which allows us to evaluate all 100+ data splits via the call \texttt{gem\_metrics outputs.json}.

\noindent \textbf{J5 - Add a new Metric}
In \texttt{gem-metrics}, each metric implements a \texttt{compute()} function and our library handles caching, parallelism, tokenization, etc. To avoid dependency conflicts, a metric can optionally specify a docker environment, as suggested by \citet{deutsch2022repro}.

\begin{minted}[fontsize=\footnotesize,autogobble,frame=single]{python}
from .texts import Predictions
from .texts import References
from .metric import ReferencedMetric

class NewMetric(ReferencedMetric):
  def _initialize(self):
    """Load models and artifacts."""
    pass

  def compute(
      self,
      cache,
      predictions: Predictions,
      references: References) -> Dict:
    """Compute the metric."""
    pass
\end{minted}

\noindent \textbf{J6 - Use Prior Results}
Comparisons to prior work often only copy reported numbers which could be computed using different evaluation parameters, and a lack of released model outputs frequently prevents a fair side-by-side comparison outside of leaderboards~\citep{gehrmann2022repairing}. To improve comparability, we add every submission to the online metrics computation to a growing corpus of model outputs which evaluation researchers can use to develop better metrics or to conduct analyses. All online submissions also appear in the result exploration tool we released with GEMv1.

\begin{figure*}[ht!]
\centering
\includegraphics[width=\textwidth]{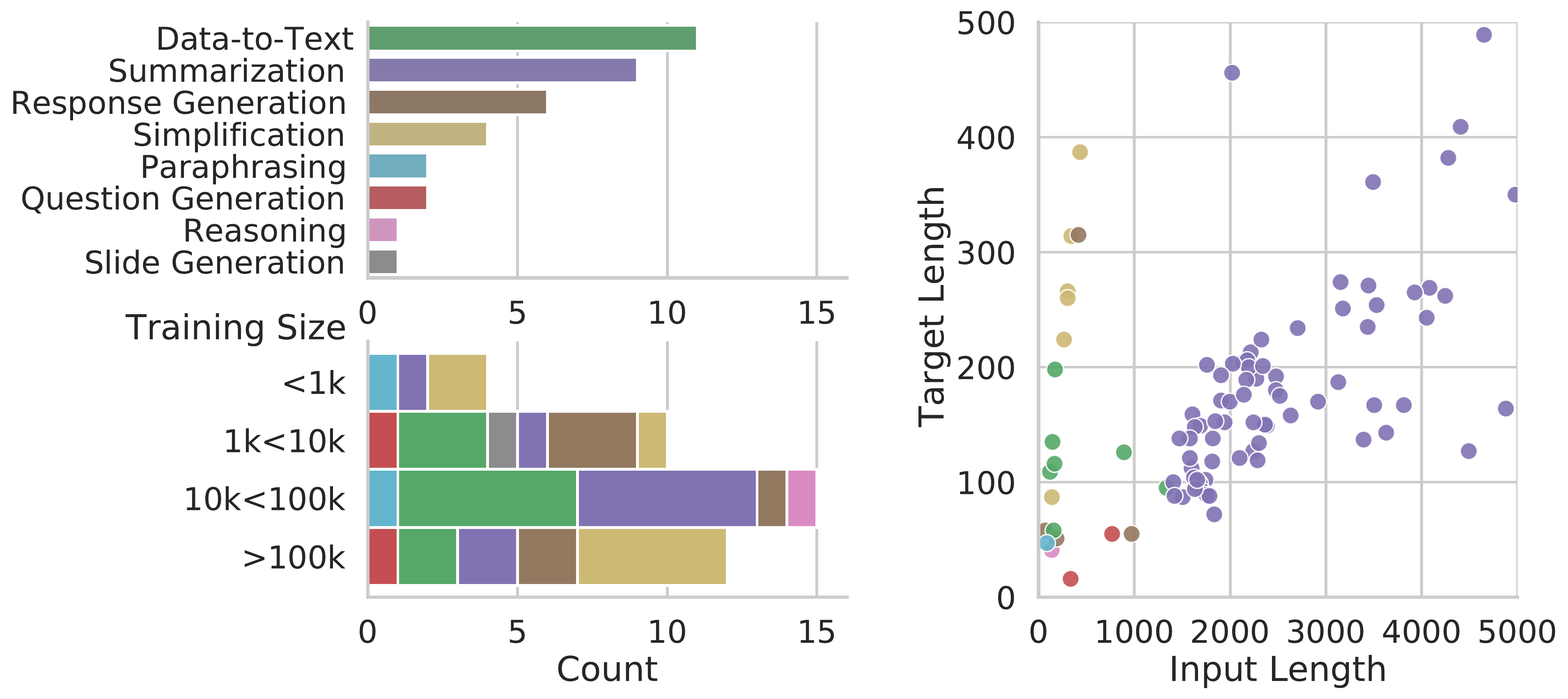}
\caption{An overview of the properties of the currently supported datasets in GEM. (Top left) A histogram of the supported task types. The most represented tasks are Data-to-Text, followed by Summarization, Response Generation, and Simplification. (Bottom Left) The frequency of different training corpus sizes for dataset configurations, broken down by their task types. While some task types are represented across all resource availability levels, some are concentrated on high resource. (Right) An overview of input and target lengths of different dataset configurations according to the mT5 tokenizer~\citep{xue-etal-2021-mt5}. Summarization tasks have input lengths of over 1,000 while all other tasks remain under 1,000 tokens. There is a lot more between-task variance in output length. Four dataset configurations are hidden due to the axis truncation. }
\label{fig:overview}
\end{figure*}

\section{Dataset Selection and Loading}
\label{sec:dataset_selection}

To identify candidate datasets for GEMv2, we followed the SuperGLUE process~\citep{DBLP:conf/nips/WangPNSMHLB19} which we already used for GEMv1 and solicited tasks to be included from the research community. Our request to suggest multilingual, challenging, and/or interesting NLG tasks led to 40 submissions. To avoid quality judgments, we imposed only three requirements to be selected: (1) dataset authors need to consent, (2) the data needs to be openly available under a permissive license, (3) the task needs to be able to be cast as a text-to-text problem. 27 new tasks were selected in addition to porting the 13 existing ones~\citep{gehrmann-etal-2021-gem}, and we also redesigned data splits for an existing task~\citep[WikiLingua,][]{ladhak-etal-2020-wikilingua}. Three of the datasets are simplification evaluation sets added to the WikiAuto loader~\citep{jiang-etal-2020-neural}, while all others have independent data loaders.

All data loaders and cards were produced as part of a month-long hackathon, and we invited the original dataset authors and GEM participants to contribute to one or more of the datasets. Afterwards, the organizers managed the ongoing maintenance. New datasets can be added on an ongoing basis, as long as the three requirements are fulfilled. GEMv2 currently supports 40 datasets, listed in Appendix~\ref{app:overview} and described in this section.

Figure~\ref{fig:overview} shows the distributions of training example count, task types, and their input and target lengths. Data-to-text and summarization are most common, followed by response generation. While data-to-text tasks are spread across resource availability categories, summarization datasets tend to be larger. While datasets vary in target length, the median input length tends to remain under 500 tokens, likely motivated by modeling limitations. Exceptions to this are summarization, with input lengths beyond what is supported by most models (e.g., WikiCatSum \citep{perez-beltrachini-etal-2019-generating} and XLSum \cite{hasan-etal-2021-xl}), and a class of data-to-text datasets with the communicative goal to generate game summaries from large sports statistic tables~\citep[e.g.,][]{hayashi-etal-2019-findings,thomson-etal-2020-sportsett,puduppully-etal-2019-data}.

We put a strong emphasis on language diversity, as prior work has found that fewer than 30\% of NLG publications (even counting evaluations on machine translation) evaluate on non-English tasks~\citep{gehrmann2022repairing}. While a lot of this focus on English can be traced to a lack of multilingual resources, many non-English NLG datasets have been released in recent years~\citep[e.g.,][]{hasan-etal-2021-xl,ladhak-etal-2020-wikilingua,mille-etal-2020-third,cahyawijaya-etal-2021-indonlg}. As shown in Table~\ref{tab:langtax}, we support languages across all resource classes in the taxonomy by \citet{joshi-etal-2020-state}. However, the focus on English is still apparent in the number of datasets supporting a particular language, shown in Table~\ref{tab:langcount}, where English is far above all other languages.
Moreover, most of the language diversity stems from the three highly multilingual datasets XLSum~\citep{hasan-etal-2021-xl}, WikiLingua~\citep{ladhak-etal-2020-wikilingua}, and data from the surface realization shared task '20~\citep{mille-etal-2020-third}. Excluding those, there are 13 datasets supporting non-English languages, 9 of which are exclusively non-English.

Of the 40 datasets, 14 have multiple configurations which can differ in task setup, languages, their encoding in romanized or original script, or domain. For example, we modified WikiLingua~\citep{ladhak-etal-2020-wikilingua} to have splits from and to any of the 18 supported languages, enabling better cross-lingual evaluations. Seventeen datasets have challenge splits, many of which were created for GEM. For example, the challenge set for the conversational weather dataset~\citep{balakrishnan-etal-2019-constrained} selects examples from the original test split with complex discourse relations.

\begin{table}[t]
\small
\begin{tabular}{lp{6cm}}
\toprule
Count & Languages \\ \midrule
1 & Amharic, Azerbaijani, Bengali, Burmese, Dutch, Gujarati, Hausa, Igbo, Javanese, Kirundi, Kyrgyz, Marathi, Nepali, Oromo, Pashto, Persian, Pidgin, Punjabi, Scottish Gaelic, Serbian, Sinhala, Somali, Sundanese, Swahili, Swedish, Tamil, Telugu, Tigrinya, Ukrainian, Urdu, Uzbek, Welsh, Yoruba \\
2 & Czech, Italian, Thai, Turkish, Vietnamese \\
3 & Arabic, Finnish, Hindi, Japanese, Korean, Portuguese \\
4 & Indonesian \\
6 & Chinese, German, Russian, Spanish \\
8 & French \\
28 & English \\ \bottomrule
\end{tabular}
\caption{The languages supported in GEMv2 and in how many of its datasets they appear.}
\label{tab:langcount}
\end{table}

\section{Data Cards}

Each dataset is accompanied by documentation about how it was created, who created it, how it should be used, and the risks in using it~\citep{bender-friedman-2018-data,gebru2018datasheets}.
Our original data documentation process~\citep{mcmillan-major-etal-2021-reusable} required filling out a markdown template following instructions in a separate guide. We analyzed the existing template and the resulting data cards under the dimensions provided in the data card playbook~\citep{pushkarna2021datacardsplaybookupdated} and identified the following improvements:
\begin{itemize}[leftmargin=*]
    \item \textbf{Accountability:} It needs to be clear who will maintain and extend the data cards when a dataset changes, when limitations of a dataset are found, or when it is deprecated~\citep{DBLP:journals/corr/abs-2111-04424}.
    \item \textbf{Utility:} The recommended evaluation process for a dataset should be prominently shown.
    \item \textbf{Quality:} We need a process to validate data card completeness and quality.
    \item \textbf{Impact \& Consequences:} It needs to be clear that we are curators, not editors, and that critiques reflect on the data, not the creators.
    \item \textbf{Risk \& Recommendations I:} We need to expand the documentation of potential PII issues.
    \item \textbf{Risk \& Recommendations II:} To help decide whether to use a dataset, the card needs to discuss differences from other datasets with similar communicative goals.
\end{itemize}

\noindent We modified our template following these insights and to be in line with the playbook approach of dividing between \textit{telescope}, \textit{periscope}, and \textit{microscope} questions based on the length of the expected answer.
We implemented this template in an \href{https://huggingface.co/spaces/GEM/DatasetCardForm}{interactive collection tool} that can create new cards or load and update existing ones. The tool shows progress bars for the overall answer status and a breakdown for each of the subsections to indicate where more content should be added.
The tool further improves the user experience by conditionally rendering questions based on prior answers, e.g., \emph{Is there a risk of PII?} $\rightarrow$ \emph{What kind of PII?}

The output of the tool is a structured json file that we convert into a simple markdown file for the data loader and an optimized web viewer and embedded in our website (Figure~\ref{fig:datacard}). The viewer presents important information at the top and splits the detailed rendering into three columns, corresponding to the telescope, periscope, and microscope split. This enables an easy navigation since high-level information can be found by focusing on the left column, moving toward the right for additional details.

\begin{table}[t]
\small
\begin{tabular}{@{}lp{6cm}}
\toprule
Tax. & Languages \\ \midrule
0 & West African Pidgin English, Sinhala \\
1 & Azerbaijani, Burmese, Gujarati, Igbo, Javanese, Kirundi, Kyrgyz, Nepali, Oromo, Pashto, Scottish Gaelic, Somali, Sundanese, Telugu, Welsh \\
2 & Amharic, Hausa, Marathi, Punjabi, Swahili, Tigrinya, Yoruba \\
3 & Bengali, Indonesian, Tamil, Thai, Ukrainian, Urdu, Uzbek \\
4 & Czech, Dutch, Finnish, Hindi, Italian, Korean, Persian, Portuguese, Russian, Serbian, Swedish, Turkish, Vietnamese \\
5 & Arabic, Chinese, English, French, German, Japanese, Spanish \\ \bottomrule
\end{tabular}
\caption{Supported languages categorized into the resource taxonomy by \citet{joshi-etal-2020-state}.}
\label{tab:langtax}
\end{table}

\begin{figure*}[ht]
\centering
\includegraphics[scale=0.30]{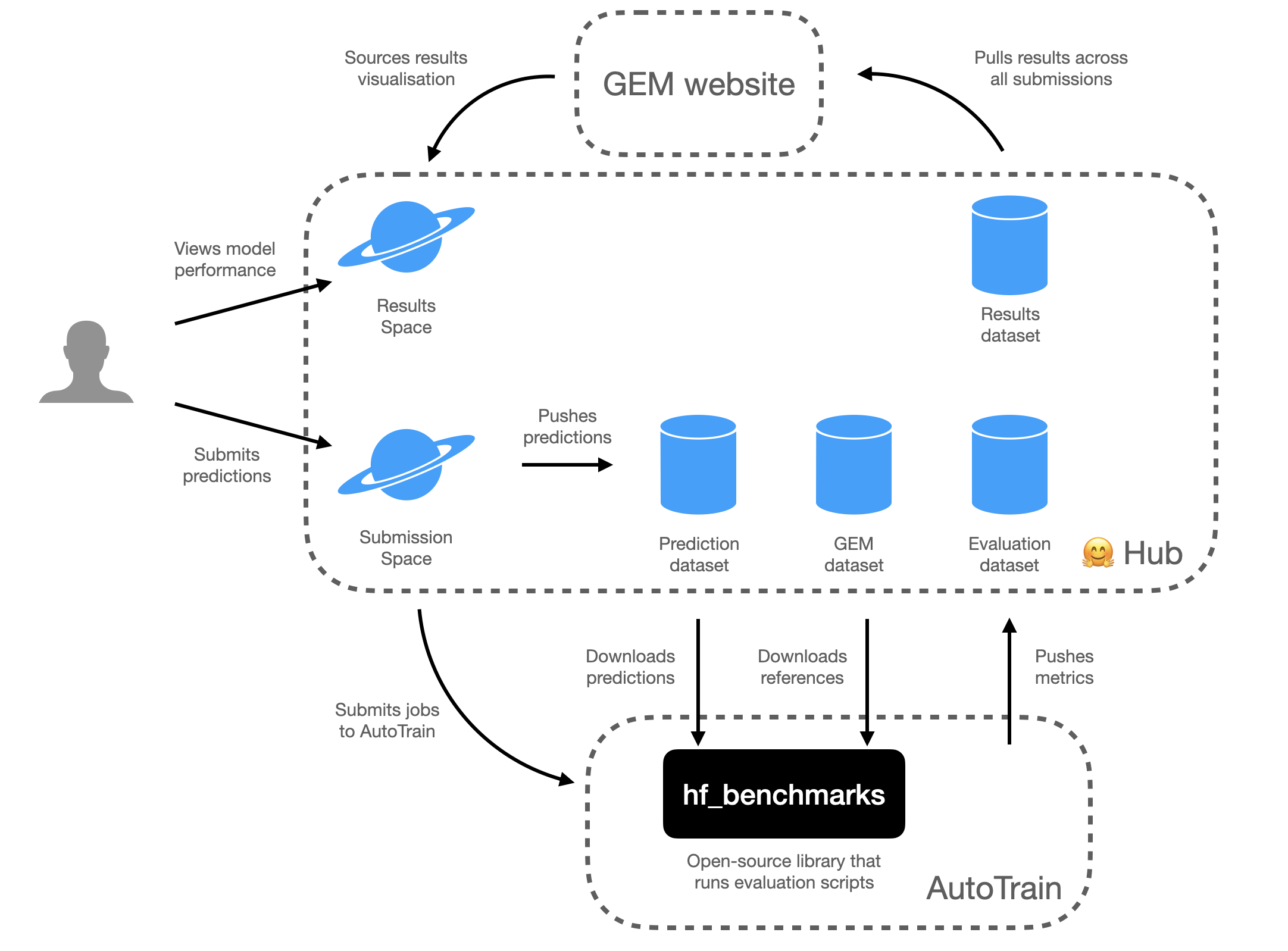}
\caption{System architecture for hosting GEM on the Hugging Face Hub}
\label{fig:architecture}
\end{figure*}

The structured format enables us to study trends in dataset construction practices beyond those shown in Section~\ref{sec:dataset_selection}. We show some exemplary statistics below, but encourage others to use the publicly available files for their investigations.
For example, 20 of the data cards report that PII is unlikely or definitely not included, while it is likely or definitely included in 10. In the free-text explanations, we find four different types of justifications for absent PII: The majority (7) stated that the data format or domain was restricted to avoid PII. Two datasets were based on public domain data (e.g., Wikipedia) and another two used fully simulated data. One response described that crowd raters were instructed to avoid any mention of PII.
We also find that multiple of the PII-likely datasets only use public domain data, indicating that there may be confusion about PII definitions.

Investigating the licensing status of our datasets, we find that the vast majority uses different variants of the Creative Commons licenses (22), 4 use the MIT license and 3 use Apache 2.0. The majority of datasets allows the unrestricted use of datasets, with 8 limiting the use to non-commercial use cases. This distribution is likely skewed due to our selection restriction to publicly available datasets.

Another typically hidden aspect is the data sourcing. Our datasets present an almost even split between automatically-, crowdworker-, and expert-created datasets, with crowdworker-created ones being slightly more common, possibly confounded if experts were hired through crowdworking platforms, as was done for SQuality~\cite{wang2022squality}.
It may thus also possible to compare which of these collection methods leads to more insightful modeling results. We follow up by asking which crowdworking platform was used and unsurprisingly, Amazon Mechanical Turk was the most frequent answer, followed by participatory experiments and other non-specified platforms.

\section{System Design}

To support the automatic evaluation of outputs, we use the Hugging Face Hub to integrate datasets, metrics, and user interfaces for GEM users to submit their outputs. The system architecture is shown in Figure~\ref{fig:architecture}, and consists of five main components:

\noindent \textbf{Spaces}  We host Streamlit applications on Spaces\footnote{\url{huggingface.co/spaces}} for the submission of predictions, downloading of results, and visualization of model performance.

\noindent \textbf{Datasets} Dataset repositories are used to host the datasets, submissions, evaluations, and results.

\noindent \textbf{AutoTrain} We use AutoTrain\footnote{\url{huggingface.co/autotrain}}, Hugging Face's AutoML platform, to run all evaluation jobs using \textbf{Hugging Face Benchmarks}, a library that defines how metrics are computed within AutoTrain.\footnote{\url{github.com/huggingface/hf_benchmarks}}

\noindent \textbf{Metrics} We use \texttt{GEM-metrics} to perform the metric computations. In addition to supporting common metrics like BLEU~\citep{papineni-etal-2002-bleu} and ROUGE~\citep{lin-2004-rouge}, the Docker integration simplifies the calculation of multiple model-based metrics like BLEURT~\citep{sellam-etal-2020-bleurt}.

On submission, a dataset repository with the model outputs is created under the \texttt{GEM-submissions} organisation on the Hugging Face Hub. In parallel, an evaluation job is triggered in AutoTrain which downloads the submission from the Hub, along with all the reference splits of the GEM datasets. These references are used to compute a wide variety of NLG metrics via GEM-metrics. The resulting metrics are then pushed to a dataset repository on the Hub, and used to source the visualization of results on the GEM website\footnote{\url{gem-benchmark.com}} and Space.\footnote{\url{huggingface.co/spaces/GEM/results}}

\section{Conclusion}
We introduce GEMv2 which aims to unify infrastructure for generation research. We propose a consistent workflow from documenting and choosing datasets to loading and evaluating on them while keeping all supported datasets and metrics compatible with each other. We demonstrate the scalability of our format by releasing the initial version with support for 40 datasets in 51 languages.
Of the supported datasets, 23 are improved through configurations, filtering, and re-splitting processes and 17 datasets have challenge sets.
Finally, we release a submission tool that computes metrics and makes model outputs available to download for evaluation researchers. Researchers who are interested in integrating their dataset are welcome to contact us for support.

\bibliography{anthology,custom}

\begin{thebibliography}{77}
\expandafter\ifx\csname natexlab\endcsname\relax\def\natexlab#1{#1}\fi

\bibitem[{Alva-Manchego et~al.(2020)Alva-Manchego, Martin, Bordes, Scarton,
  Sagot, and Specia}]{alva-manchego-etal-2020-asset}
Fernando Alva-Manchego, Louis Martin, Antoine Bordes, Carolina Scarton,
  Beno{\^\i}t Sagot, and Lucia Specia. 2020.
\newblock \href {https://doi.org/10.18653/v1/2020.acl-main.424} {{ASSET}: {A}
  dataset for tuning and evaluation of sentence simplification models with
  multiple rewriting transformations}.
\newblock In \emph{Proceedings of the 58th Annual Meeting of the Association
  for Computational Linguistics}, pages 4668--4679, Online. Association for
  Computational Linguistics.

\bibitem[{Balakrishnan et~al.(2019)Balakrishnan, Rao, Upasani, White, and
  Subba}]{balakrishnan-etal-2019-constrained}
Anusha Balakrishnan, Jinfeng Rao, Kartikeya Upasani, Michael White, and Rajen
  Subba. 2019.
\newblock \href {https://doi.org/10.18653/v1/P19-1080} {Constrained decoding
  for neural {NLG} from compositional representations in task-oriented
  dialogue}.
\newblock In \emph{Proceedings of the 57th Annual Meeting of the Association
  for Computational Linguistics}, pages 831--844, Florence, Italy. Association
  for Computational Linguistics.

\bibitem[{Belz et~al.(2021)Belz, Agarwal, Shimorina, and
  Reiter}]{belz-etal-2021-systematic}
Anya Belz, Shubham Agarwal, Anastasia Shimorina, and Ehud Reiter. 2021.
\newblock \href {https://doi.org/10.18653/v1/2021.eacl-main.29} {A systematic
  review of reproducibility research in natural language processing}.
\newblock In \emph{Proceedings of the 16th Conference of the European Chapter
  of the Association for Computational Linguistics: Main Volume}, pages
  381--393, Online. Association for Computational Linguistics.

\bibitem[{Bender and Friedman(2018)}]{bender-friedman-2018-data}
Emily~M. Bender and Batya Friedman. 2018.
\newblock \href {https://doi.org/10.1162/tacl_a_00041} {Data statements for
  natural language processing: Toward mitigating system bias and enabling
  better science}.
\newblock \emph{Transactions of the Association for Computational Linguistics},
  6:587--604.

\bibitem[{Bhagavatula et~al.(2020)Bhagavatula, Bras, Malaviya, Sakaguchi,
  Holtzman, Rashkin, Downey, tau Yih, and Choi}]{Bhagavatula2020Abductive}
Chandra Bhagavatula, Ronan~Le Bras, Chaitanya Malaviya, Keisuke Sakaguchi, Ari
  Holtzman, Hannah Rashkin, Doug Downey, Wen tau Yih, and Yejin Choi. 2020.
\newblock \href {https://openreview.net/forum?id=Byg1v1HKDB} {Abductive
  commonsense reasoning}.
\newblock In \emph{International Conference on Learning Representations}.

\bibitem[{Bowman and Dahl(2021)}]{bowman-dahl-2021-will}
Samuel~R. Bowman and George Dahl. 2021.
\newblock \href {https://doi.org/10.18653/v1/2021.naacl-main.385} {What will it
  take to fix benchmarking in natural language understanding?}
\newblock In \emph{Proceedings of the 2021 Conference of the North American
  Chapter of the Association for Computational Linguistics: Human Language
  Technologies}, pages 4843--4855, Online. Association for Computational
  Linguistics.

\bibitem[{Byrne et~al.(2021)Byrne, Krishnamoorthi, Ganesh, and
  Kale}]{byrne-etal-2021-tickettalk}
Bill Byrne, Karthik Krishnamoorthi, Saravanan Ganesh, and Mihir Kale. 2021.
\newblock \href {https://doi.org/10.18653/v1/2021.acl-long.55} {{T}icket{T}alk:
  Toward human-level performance with end-to-end, transaction-based dialog
  systems}.
\newblock In \emph{Proceedings of the 59th Annual Meeting of the Association
  for Computational Linguistics and the 11th International Joint Conference on
  Natural Language Processing (Volume 1: Long Papers)}, pages 671--680, Online.
  Association for Computational Linguistics.

\bibitem[{Byrne et~al.(2019)Byrne, Krishnamoorthi, Sankar, Neelakantan,
  Goodrich, Duckworth, Yavuz, Dubey, Kim, and
  Cedilnik}]{byrne-etal-2019-taskmaster}
Bill Byrne, Karthik Krishnamoorthi, Chinnadhurai Sankar, Arvind Neelakantan,
  Ben Goodrich, Daniel Duckworth, Semih Yavuz, Amit Dubey, Kyu-Young Kim, and
  Andy Cedilnik. 2019.
\newblock \href {https://doi.org/10.18653/v1/D19-1459} {Taskmaster-1: Toward a
  realistic and diverse dialog dataset}.
\newblock In \emph{Proceedings of the 2019 Conference on Empirical Methods in
  Natural Language Processing and the 9th International Joint Conference on
  Natural Language Processing (EMNLP-IJCNLP)}, pages 4516--4525, Hong Kong,
  China. Association for Computational Linguistics.

\bibitem[{Cahyawijaya et~al.(2021)Cahyawijaya, Winata, Wilie, Vincentio, Li,
  Kuncoro, Ruder, Lim, Bahar, Khodra, Purwarianti, and
  Fung}]{cahyawijaya-etal-2021-indonlg}
Samuel Cahyawijaya, Genta~Indra Winata, Bryan Wilie, Karissa Vincentio,
  Xiaohong Li, Adhiguna Kuncoro, Sebastian Ruder, Zhi~Yuan Lim, Syafri Bahar,
  Masayu Khodra, Ayu Purwarianti, and Pascale Fung. 2021.
\newblock \href {https://doi.org/10.18653/v1/2021.emnlp-main.699} {{I}ndo{NLG}:
  Benchmark and resources for evaluating {I}ndonesian natural language
  generation}.
\newblock In \emph{Proceedings of the 2021 Conference on Empirical Methods in
  Natural Language Processing}, pages 8875--8898, Online and Punta Cana,
  Dominican Republic. Association for Computational Linguistics.

\bibitem[{Corry et~al.(2021)Corry, Sridharan, Luccioni, Ananny, Schultz, and
  Crawford}]{DBLP:journals/corr/abs-2111-04424}
Frances Corry, Hamsini Sridharan, Alexandra~Sasha Luccioni, Mike Ananny, Jason
  Schultz, and Kate Crawford. 2021.
\newblock \href {http://arxiv.org/abs/2111.04424} {The problem of zombie
  datasets: {A} framework for deprecating datasets}.
\newblock \emph{CoRR}, abs/2111.04424.

\bibitem[{Creutz(2018)}]{creutz:lrec2018}
Mathias Creutz. 2018.
\newblock \href {http://www.lrec-conf.org/proceedings/lrec2018/pdf/131.pdf}
  {Open subtitles paraphrase corpus for six languages}.
\newblock In \emph{Proceedings of the 11th edition of the Language Resources
  and Evaluation Conference (LREC 2018)}, Miyazaki, Japan. European Language
  Resources Association (ELRA).

\bibitem[{Dehghani et~al.(2021)Dehghani, Tay, Gritsenko, Zhao, Houlsby, Diaz,
  Metzler, and Vinyals}]{DBLP:journals/corr/abs-2107-07002}
Mostafa Dehghani, Yi~Tay, Alexey~A. Gritsenko, Zhe Zhao, Neil Houlsby, Fernando
  Diaz, Donald Metzler, and Oriol Vinyals. 2021.
\newblock \href {http://arxiv.org/abs/2107.07002} {The benchmark lottery}.
\newblock \emph{CoRR}, abs/2107.07002.

\bibitem[{Deutsch and Roth(2022)}]{deutsch2022repro}
Daniel Deutsch and Dan Roth. 2022.
\newblock \href {https://arxiv.org/abs/2204.13848} {{Repro: An Open-Source
  Library for Improving the Reproducibility and Usability of Publicly Available
  Research Code}}.
\newblock \emph{ArXiv}, abs/2204.13848.

\bibitem[{Devaraj et~al.(2021)Devaraj, Marshall, Wallace, and
  Li}]{devaraj-etal-2021-paragraph}
Ashwin Devaraj, Iain Marshall, Byron Wallace, and Junyi~Jessy Li. 2021.
\newblock \href {https://doi.org/10.18653/v1/2021.naacl-main.395}
  {Paragraph-level simplification of medical texts}.
\newblock In \emph{Proceedings of the 2021 Conference of the North American
  Chapter of the Association for Computational Linguistics: Human Language
  Technologies}, pages 4972--4984, Online. Association for Computational
  Linguistics.

\bibitem[{Dhole et~al.(2021)Dhole, Gangal, Gehrmann, Gupta, Li, Mahamood,
  Mahendiran, Mille, Srivastava, Tan et~al.}]{dhole2021nl}
Kaustubh~D Dhole, Varun Gangal, Sebastian Gehrmann, Aadesh Gupta, Zhenhao Li,
  Saad Mahamood, Abinaya Mahendiran, Simon Mille, Ashish Srivastava, Samson
  Tan, et~al. 2021.
\newblock Nl-augmenter: A framework for task-sensitive natural language
  augmentation.
\newblock \emph{arXiv preprint arXiv:2112.02721}.

\bibitem[{Du{\v{s}}ek et~al.(2019)Du{\v{s}}ek, Howcroft, and
  Rieser}]{dusek-etal-2019-semantic}
Ond{\v{r}}ej Du{\v{s}}ek, David~M. Howcroft, and Verena Rieser. 2019.
\newblock \href {https://doi.org/10.18653/v1/W19-8652} {Semantic noise matters
  for neural natural language generation}.
\newblock In \emph{Proceedings of the 12th International Conference on Natural
  Language Generation}, pages 421--426, Tokyo, Japan. Association for
  Computational Linguistics.

\bibitem[{Du{\v{s}}ek and
  Jur{\v{c}}{\'\i}{\v{c}}ek(2019)}]{dusek-jurcicek-2019-neural}
Ond{\v{r}}ej Du{\v{s}}ek and Filip Jur{\v{c}}{\'\i}{\v{c}}ek. 2019.
\newblock \href {https://doi.org/10.18653/v1/W19-8670} {Neural generation for
  {C}zech: Data and baselines}.
\newblock In \emph{Proceedings of the 12th International Conference on Natural
  Language Generation}, pages 563--574, Tokyo, Japan. Association for
  Computational Linguistics.

\bibitem[{Dušek et~al.(2020)Dušek, Novikova, and Rieser}]{DUSEK2020123}
Ondřej Dušek, Jekaterina Novikova, and Verena Rieser. 2020.
\newblock \href {https://doi.org/https://doi.org/10.1016/j.csl.2019.06.009}
  {Evaluating the state-of-the-art of end-to-end natural language generation:
  The e2e nlg challenge}.
\newblock \emph{Computer Speech \& Language}, 59:123--156.

\bibitem[{Ethayarajh and Jurafsky(2020)}]{ethayarajh-jurafsky-2020-utility}
Kawin Ethayarajh and Dan Jurafsky. 2020.
\newblock \href {https://doi.org/10.18653/v1/2020.emnlp-main.393} {Utility is
  in the eye of the user: A critique of {NLP} leaderboards}.
\newblock In \emph{Proceedings of the 2020 Conference on Empirical Methods in
  Natural Language Processing (EMNLP)}, pages 4846--4853, Online. Association
  for Computational Linguistics.

\bibitem[{Gardent et~al.(2017{\natexlab{a}})Gardent, Shimorina, Narayan, and
  Perez-Beltrachini}]{gardent-etal-2017-creating}
Claire Gardent, Anastasia Shimorina, Shashi Narayan, and Laura
  Perez-Beltrachini. 2017{\natexlab{a}}.
\newblock \href {https://doi.org/10.18653/v1/P17-1017} {Creating training
  corpora for {NLG} micro-planners}.
\newblock In \emph{Proceedings of the 55th Annual Meeting of the Association
  for Computational Linguistics (Volume 1: Long Papers)}, pages 179--188,
  Vancouver, Canada. Association for Computational Linguistics.

\bibitem[{Gardent et~al.(2017{\natexlab{b}})Gardent, Shimorina, Narayan, and
  Perez-Beltrachini}]{gardent-etal-2017-webnlg}
Claire Gardent, Anastasia Shimorina, Shashi Narayan, and Laura
  Perez-Beltrachini. 2017{\natexlab{b}}.
\newblock \href {https://doi.org/10.18653/v1/W17-3518} {The {W}eb{NLG}
  challenge: Generating text from {RDF} data}.
\newblock In \emph{Proceedings of the 10th International Conference on Natural
  Language Generation}, pages 124--133, Santiago de Compostela, Spain.
  Association for Computational Linguistics.

\bibitem[{Gebru et~al.(2018)Gebru, Morgenstern, Vecchione, Vaughan, Wallach,
  Daum{\'e}~III, and Crawford}]{gebru2018datasheets}
Timnit Gebru, Jamie Morgenstern, Briana Vecchione, Jennifer~Wortman Vaughan,
  Hanna Wallach, Hal Daum{\'e}~III, and Kate Crawford. 2018.
\newblock \href {http://arxiv.org/abs/1803.09010} {Datasheets for datasets}.
\newblock In \emph{Proceedings of the Fifth Workshop on Fairness,
  Accountability, and Transparency in Machine Learning}, Stockholm, Sweden.

\bibitem[{Gehrmann et~al.(2021)Gehrmann, Adewumi, Aggarwal, Ammanamanchi,
  Aremu, Bosselut, Chandu, Clinciu, Das, Dhole, Du, Durmus, Du{\v{s}}ek,
  Emezue, Gangal, Garbacea, Hashimoto, Hou, Jernite, Jhamtani, Ji, Jolly, Kale,
  Kumar, Ladhak, Madaan, Maddela, Mahajan, Mahamood, Majumder, Martins,
  McMillan-Major, Mille, van Miltenburg, Nadeem, Narayan, Nikolaev,
  Niyongabo~Rubungo, Osei, Parikh, Perez-Beltrachini, Rao, Raunak, Rodriguez,
  Santhanam, Sedoc, Sellam, Shaikh, Shimorina, Sobrevilla~Cabezudo, Strobelt,
  Subramani, Xu, Yang, Yerukola, and Zhou}]{gehrmann-etal-2021-gem}
Sebastian Gehrmann, Tosin Adewumi, Karmanya Aggarwal, Pawan~Sasanka
  Ammanamanchi, Anuoluwapo Aremu, Antoine Bosselut, Khyathi~Raghavi Chandu,
  Miruna-Adriana Clinciu, Dipanjan Das, Kaustubh Dhole, Wanyu Du, Esin Durmus,
  Ond{\v{r}}ej Du{\v{s}}ek, Chris~Chinenye Emezue, Varun Gangal, Cristina
  Garbacea, Tatsunori Hashimoto, Yufang Hou, Yacine Jernite, Harsh Jhamtani,
  Yangfeng Ji, Shailza Jolly, Mihir Kale, Dhruv Kumar, Faisal Ladhak, Aman
  Madaan, Mounica Maddela, Khyati Mahajan, Saad Mahamood, Bodhisattwa~Prasad
  Majumder, Pedro~Henrique Martins, Angelina McMillan-Major, Simon Mille, Emiel
  van Miltenburg, Moin Nadeem, Shashi Narayan, Vitaly Nikolaev, Andre
  Niyongabo~Rubungo, Salomey Osei, Ankur Parikh, Laura Perez-Beltrachini,
  Niranjan~Ramesh Rao, Vikas Raunak, Juan~Diego Rodriguez, Sashank Santhanam,
  Jo{\~a}o Sedoc, Thibault Sellam, Samira Shaikh, Anastasia Shimorina,
  Marco~Antonio Sobrevilla~Cabezudo, Hendrik Strobelt, Nishant Subramani, Wei
  Xu, Diyi Yang, Akhila Yerukola, and Jiawei Zhou. 2021.
\newblock \href {https://doi.org/10.18653/v1/2021.gem-1.10} {The {GEM}
  benchmark: Natural language generation, its evaluation and metrics}.
\newblock In \emph{Proceedings of the 1st Workshop on Natural Language
  Generation, Evaluation, and Metrics (GEM 2021)}, pages 96--120, Online.
  Association for Computational Linguistics.

\bibitem[{Gehrmann et~al.(2022)Gehrmann, Clark, and
  Sellam}]{gehrmann2022repairing}
Sebastian Gehrmann, Elizabeth Clark, and Thibault Sellam. 2022.
\newblock \href {http://arxiv.org/abs/2202.06935} {Repairing the cracked
  foundation: {A} survey of obstacles in evaluation practices for generated
  text}.
\newblock \emph{CoRR}, abs/2202.06935.

\bibitem[{Hasan et~al.(2021)Hasan, Bhattacharjee, Islam, Mubasshir, Li, Kang,
  Rahman, and Shahriyar}]{hasan-etal-2021-xl}
Tahmid Hasan, Abhik Bhattacharjee, Md.~Saiful Islam, Kazi Mubasshir, Yuan-Fang
  Li, Yong-Bin Kang, M.~Sohel Rahman, and Rifat Shahriyar. 2021.
\newblock \href {https://doi.org/10.18653/v1/2021.findings-acl.413} {{XL}-sum:
  Large-scale multilingual abstractive summarization for 44 languages}.
\newblock In \emph{Findings of the Association for Computational Linguistics:
  ACL-IJCNLP 2021}, pages 4693--4703, Online. Association for Computational
  Linguistics.

\bibitem[{Hayashi et~al.(2019)Hayashi, Oda, Birch, Konstas, Finch, Luong,
  Neubig, and Sudoh}]{hayashi-etal-2019-findings}
Hiroaki Hayashi, Yusuke Oda, Alexandra Birch, Ioannis Konstas, Andrew Finch,
  Minh-Thang Luong, Graham Neubig, and Katsuhito Sudoh. 2019.
\newblock \href {https://doi.org/10.18653/v1/D19-5601} {Findings of the third
  workshop on neural generation and translation}.
\newblock In \emph{Proceedings of the 3rd Workshop on Neural Generation and
  Translation}, pages 1--14, Hong Kong. Association for Computational
  Linguistics.

\bibitem[{Howcroft et~al.(2020)Howcroft, Belz, Clinciu, Gkatzia, Hasan,
  Mahamood, Mille, van Miltenburg, Santhanam, and
  Rieser}]{howcroft-etal-2020-twenty}
David~M. Howcroft, Anya Belz, Miruna-Adriana Clinciu, Dimitra Gkatzia, Sadid~A.
  Hasan, Saad Mahamood, Simon Mille, Emiel van Miltenburg, Sashank Santhanam,
  and Verena Rieser. 2020.
\newblock \href {https://aclanthology.org/2020.inlg-1.23} {Twenty years of
  confusion in human evaluation: {NLG} needs evaluation sheets and standardised
  definitions}.
\newblock In \emph{Proceedings of the 13th International Conference on Natural
  Language Generation}, pages 169--182, Dublin, Ireland. Association for
  Computational Linguistics.

\bibitem[{Jiang et~al.(2020)Jiang, Maddela, Lan, Zhong, and
  Xu}]{jiang-etal-2020-neural}
Chao Jiang, Mounica Maddela, Wuwei Lan, Yang Zhong, and Wei Xu. 2020.
\newblock \href {https://doi.org/10.18653/v1/2020.acl-main.709} {Neural {CRF}
  model for sentence alignment in text simplification}.
\newblock In \emph{Proceedings of the 58th Annual Meeting of the Association
  for Computational Linguistics}, pages 7943--7960, Online. Association for
  Computational Linguistics.

\bibitem[{Joshi et~al.(2020)Joshi, Santy, Budhiraja, Bali, and
  Choudhury}]{joshi-etal-2020-state}
Pratik Joshi, Sebastin Santy, Amar Budhiraja, Kalika Bali, and Monojit
  Choudhury. 2020.
\newblock \href {https://doi.org/10.18653/v1/2020.acl-main.560} {The state and
  fate of linguistic diversity and inclusion in the {NLP} world}.
\newblock In \emph{Proceedings of the 58th Annual Meeting of the Association
  for Computational Linguistics}, pages 6282--6293, Online. Association for
  Computational Linguistics.

\bibitem[{Juraska et~al.(2019)Juraska, Bowden, and
  Walker}]{juraska-etal-2019-viggo}
Juraj Juraska, Kevin Bowden, and Marilyn Walker. 2019.
\newblock \href {https://doi.org/10.18653/v1/W19-8623} {{V}i{GGO}: A video game
  corpus for data-to-text generation in open-domain conversation}.
\newblock In \emph{Proceedings of the 12th International Conference on Natural
  Language Generation}, pages 164--172, Tokyo, Japan. Association for
  Computational Linguistics.

\bibitem[{Kale and Rastogi(2020)}]{kale-rastogi-2020-text}
Mihir Kale and Abhinav Rastogi. 2020.
\newblock \href {https://aclanthology.org/2020.inlg-1.14} {Text-to-text
  pre-training for data-to-text tasks}.
\newblock In \emph{Proceedings of the 13th International Conference on Natural
  Language Generation}, pages 97--102, Dublin, Ireland. Association for
  Computational Linguistics.

\bibitem[{Kamal~Eddine et~al.(2021)Kamal~Eddine, Tixier, and
  Vazirgiannis}]{kamal-eddine-etal-2021-barthez}
Moussa Kamal~Eddine, Antoine Tixier, and Michalis Vazirgiannis. 2021.
\newblock \href {https://doi.org/10.18653/v1/2021.emnlp-main.740} {{BART}hez: a
  skilled pretrained {F}rench sequence-to-sequence model}.
\newblock In \emph{Proceedings of the 2021 Conference on Empirical Methods in
  Natural Language Processing}, pages 9369--9390, Online and Punta Cana,
  Dominican Republic. Association for Computational Linguistics.

\bibitem[{Kanerva et~al.(2021)Kanerva, Ginter, Chang, Rastas, Skantsi,
  Kilpel{\"a}inen, Kupari, Saarni, Sev{\'o}n, and
  Tarkka}]{kanerva-etal-2021-finnish}
Jenna Kanerva, Filip Ginter, Li-Hsin Chang, Iiro Rastas, Valtteri Skantsi,
  Jemina Kilpel{\"a}inen, Hanna-Mari Kupari, Jenna Saarni, Maija Sev{\'o}n, and
  Otto Tarkka. 2021.
\newblock \href {https://aclanthology.org/2021.nodalida-main.29} {{F}innish
  paraphrase corpus}.
\newblock In \emph{Proceedings of the 23rd Nordic Conference on Computational
  Linguistics (NoDaLiDa)}, pages 288--298, Reykjavik, Iceland (Online).
  Link{\"o}ping University Electronic Press, Sweden.

\bibitem[{Kanerva et~al.(2020)Kanerva, Ginter, and
  Pyysalo}]{kanerva-etal-2020-turku}
Jenna Kanerva, Filip Ginter, and Sampo Pyysalo. 2020.
\newblock \href {https://doi.org/10.18653/v1/2020.iwpt-1.17} {{T}urku enhanced
  parser pipeline: From raw text to enhanced graphs in the {IWPT} 2020 shared
  task}.
\newblock In \emph{Proceedings of the 16th International Conference on Parsing
  Technologies and the IWPT 2020 Shared Task on Parsing into Enhanced Universal
  Dependencies}, pages 162--173, Online. Association for Computational
  Linguistics.

\bibitem[{Kanerva et~al.(2019)Kanerva, R{\"o}nnqvist, Kekki, Salakoski, and
  Ginter}]{kanerva-etal-2019-template}
Jenna Kanerva, Samuel R{\"o}nnqvist, Riina Kekki, Tapio Salakoski, and Filip
  Ginter. 2019.
\newblock \href {https://aclanthology.org/W19-6125} {Template-free data-to-text
  generation of {F}innish sports news}.
\newblock In \emph{Proceedings of the 22nd Nordic Conference on Computational
  Linguistics}, pages 242--252, Turku, Finland. Link{\"o}ping University
  Electronic Press.

\bibitem[{Kasai et~al.(2022)Kasai, Sakaguchi, Bras, Dunagan, Morrison, Fabbri,
  Choi, and Smith}]{Kasai2021BidimensionalLG}
Jungo Kasai, Keisuke Sakaguchi, Ronan~Le Bras, Lavinia Dunagan, Jacob Morrison,
  Alexander~R. Fabbri, Yejin Choi, and Noah~A. Smith. 2022.
\newblock Bidimensional leaderboards: Generate and evaluate language hand in
  hand.
\newblock In \emph{Proceedings of the 2022 Conference of the North {A}merican
  Chapter of the Association for Computational Linguistics}, Seattle,
  Washington. Association for Computational Linguistics.

\bibitem[{Kim et~al.(2021{\natexlab{a}})Kim, Maddela, Kriz, Xu, and
  Callison-Burch}]{kim-etal-2021-bisect}
Joongwon Kim, Mounica Maddela, Reno Kriz, Wei Xu, and Chris Callison-Burch.
  2021{\natexlab{a}}.
\newblock \href {https://doi.org/10.18653/v1/2021.emnlp-main.500} {{B}i{SECT}:
  Learning to split and rephrase sentences with bitexts}.
\newblock In \emph{Proceedings of the 2021 Conference on Empirical Methods in
  Natural Language Processing}, pages 6193--6209, Online and Punta Cana,
  Dominican Republic. Association for Computational Linguistics.

\bibitem[{Kim et~al.(2021{\natexlab{b}})Kim, Liu, Jin, Papangelis,
  Gopalakrishnan, Hedayatnia, and Hakkani-T{\"u}r}]{Kim2021HowRR}
Seokhwan Kim, Yang Liu, Di~Jin, Alexandros Papangelis, Karthik Gopalakrishnan,
  Behnam Hedayatnia, and Dilek~Z. Hakkani-T{\"u}r. 2021{\natexlab{b}}.
\newblock “how robust r u?”: Evaluating task-oriented dialogue systems on
  spoken conversations.
\newblock \emph{2021 IEEE Automatic Speech Recognition and Understanding
  Workshop (ASRU)}, pages 1147--1154.

\bibitem[{Ladhak et~al.(2020)Ladhak, Durmus, Cardie, and
  McKeown}]{ladhak-etal-2020-wikilingua}
Faisal Ladhak, Esin Durmus, Claire Cardie, and Kathleen McKeown. 2020.
\newblock \href {https://doi.org/10.18653/v1/2020.findings-emnlp.360}
  {{W}iki{L}ingua: A new benchmark dataset for cross-lingual abstractive
  summarization}.
\newblock In \emph{Findings of the Association for Computational Linguistics:
  EMNLP 2020}, pages 4034--4048, Online. Association for Computational
  Linguistics.

\bibitem[{Lhoest et~al.(2021)Lhoest, Villanova~del Moral, Jernite, Thakur, von
  Platen, Patil, Chaumond, Drame, Plu, Tunstall, Davison, {\v{S}}a{\v{s}}ko,
  Chhablani, Malik, Brandeis, Le~Scao, Sanh, Xu, Patry, McMillan-Major, Schmid,
  Gugger, Delangue, Matussi{\`e}re, Debut, Bekman, Cistac, Goehringer, Mustar,
  Lagunas, Rush, and Wolf}]{lhoest-etal-2021-datasets}
Quentin Lhoest, Albert Villanova~del Moral, Yacine Jernite, Abhishek Thakur,
  Patrick von Platen, Suraj Patil, Julien Chaumond, Mariama Drame, Julien Plu,
  Lewis Tunstall, Joe Davison, Mario {\v{S}}a{\v{s}}ko, Gunjan Chhablani,
  Bhavitvya Malik, Simon Brandeis, Teven Le~Scao, Victor Sanh, Canwen Xu,
  Nicolas Patry, Angelina McMillan-Major, Philipp Schmid, Sylvain Gugger,
  Cl{\'e}ment Delangue, Th{\'e}o Matussi{\`e}re, Lysandre Debut, Stas Bekman,
  Pierric Cistac, Thibault Goehringer, Victor Mustar, Fran{\c{c}}ois Lagunas,
  Alexander Rush, and Thomas Wolf. 2021.
\newblock \href {https://doi.org/10.18653/v1/2021.emnlp-demo.21} {Datasets: A
  community library for natural language processing}.
\newblock In \emph{Proceedings of the 2021 Conference on Empirical Methods in
  Natural Language Processing: System Demonstrations}, pages 175--184, Online
  and Punta Cana, Dominican Republic. Association for Computational
  Linguistics.

\bibitem[{Lin et~al.(2020)Lin, Zhou, Shen, Zhou, Bhagavatula, Choi, and
  Ren}]{lin-etal-2020-commongen}
Bill~Yuchen Lin, Wangchunshu Zhou, Ming Shen, Pei Zhou, Chandra Bhagavatula,
  Yejin Choi, and Xiang Ren. 2020.
\newblock \href {https://doi.org/10.18653/v1/2020.findings-emnlp.165}
  {{C}ommon{G}en: A constrained text generation challenge for generative
  commonsense reasoning}.
\newblock In \emph{Findings of the Association for Computational Linguistics:
  EMNLP 2020}, pages 1823--1840, Online. Association for Computational
  Linguistics.

\bibitem[{Lin(2004)}]{lin-2004-rouge}
Chin-Yew Lin. 2004.
\newblock \href {https://aclanthology.org/W04-1013} {{ROUGE}: A package for
  automatic evaluation of summaries}.
\newblock In \emph{Text Summarization Branches Out}, pages 74--81, Barcelona,
  Spain. Association for Computational Linguistics.

\bibitem[{McMillan-Major et~al.(2021)McMillan-Major, Osei, Rodriguez,
  Ammanamanchi, Gehrmann, and Jernite}]{mcmillan-major-etal-2021-reusable}
Angelina McMillan-Major, Salomey Osei, Juan~Diego Rodriguez, Pawan~Sasanka
  Ammanamanchi, Sebastian Gehrmann, and Yacine Jernite. 2021.
\newblock \href {https://doi.org/10.18653/v1/2021.gem-1.11} {Reusable templates
  and guides for documenting datasets and models for natural language
  processing and generation: A case study of the {H}ugging{F}ace and {GEM} data
  and model cards}.
\newblock In \emph{Proceedings of the 1st Workshop on Natural Language
  Generation, Evaluation, and Metrics (GEM 2021)}, pages 121--135, Online.
  Association for Computational Linguistics.

\bibitem[{Mille et~al.(2020)Mille, Belz, Bohnet, Castro~Ferreira, Graham, and
  Wanner}]{mille-etal-2020-third}
Simon Mille, Anya Belz, Bernd Bohnet, Thiago Castro~Ferreira, Yvette Graham,
  and Leo Wanner. 2020.
\newblock \href {https://aclanthology.org/2020.msr-1.1} {The third multilingual
  surface realisation shared task ({SR}{'}20): Overview and evaluation
  results}.
\newblock In \emph{Proceedings of the Third Workshop on Multilingual Surface
  Realisation}, pages 1--20, Barcelona, Spain (Online). Association for
  Computational Linguistics.

\bibitem[{Mille et~al.(2021)Mille, Dhole, Mahamood, Perez-Beltrachini, Gangal,
  Kale, van Miltenburg, and Gehrmann}]{mille2021automatic}
Simon Mille, Kaustubh Dhole, Saad Mahamood, Laura Perez-Beltrachini, Varun
  Gangal, Mihir Kale, Emiel van Miltenburg, and Sebastian Gehrmann. 2021.
\newblock \href {https://openreview.net/forum?id=CSi1eu_2q96} {Automatic
  construction of evaluation suites for natural language generation datasets}.
\newblock In \emph{Thirty-fifth Conference on Neural Information Processing
  Systems Datasets and Benchmarks Track (Round 1)}.

\bibitem[{Nan et~al.(2021)Nan, Radev, Zhang, Rau, Sivaprasad, Hsieh, Tang,
  Vyas, Verma, Krishna, Liu, Irwanto, Pan, Rahman, Zaidi, Mutuma, Tarabar,
  Gupta, Yu, Tan, Lin, Xiong, Socher, and Rajani}]{nan-etal-2021-dart}
Linyong Nan, Dragomir Radev, Rui Zhang, Amrit Rau, Abhinand Sivaprasad,
  Chiachun Hsieh, Xiangru Tang, Aadit Vyas, Neha Verma, Pranav Krishna,
  Yangxiaokang Liu, Nadia Irwanto, Jessica Pan, Faiaz Rahman, Ahmad Zaidi,
  Mutethia Mutuma, Yasin Tarabar, Ankit Gupta, Tao Yu, Yi~Chern Tan,
  Xi~Victoria Lin, Caiming Xiong, Richard Socher, and Nazneen~Fatema Rajani.
  2021.
\newblock \href {https://doi.org/10.18653/v1/2021.naacl-main.37} {{DART}:
  Open-domain structured data record to text generation}.
\newblock In \emph{Proceedings of the 2021 Conference of the North American
  Chapter of the Association for Computational Linguistics: Human Language
  Technologies}, pages 432--447, Online. Association for Computational
  Linguistics.

\bibitem[{Narayan et~al.(2018)Narayan, Cohen, and
  Lapata}]{narayan-etal-2018-dont}
Shashi Narayan, Shay~B. Cohen, and Mirella Lapata. 2018.
\newblock \href {https://doi.org/10.18653/v1/D18-1206} {Don{'}t give me the
  details, just the summary! topic-aware convolutional neural networks for
  extreme summarization}.
\newblock In \emph{Proceedings of the 2018 Conference on Empirical Methods in
  Natural Language Processing}, pages 1797--1807, Brussels, Belgium.
  Association for Computational Linguistics.

\bibitem[{Novikova et~al.(2017)Novikova, Du{\v{s}}ek, and
  Rieser}]{novikova-etal-2017-e2e}
Jekaterina Novikova, Ond{\v{r}}ej Du{\v{s}}ek, and Verena Rieser. 2017.
\newblock \href {https://doi.org/10.18653/v1/W17-5525} {The {E}2{E} dataset:
  New challenges for end-to-end generation}.
\newblock In \emph{Proceedings of the 18th Annual {SIG}dial Meeting on
  Discourse and Dialogue}, pages 201--206, Saarbr{\"u}cken, Germany.
  Association for Computational Linguistics.

\bibitem[{Papineni et~al.(2002)Papineni, Roukos, Ward, and
  Zhu}]{papineni-etal-2002-bleu}
Kishore Papineni, Salim Roukos, Todd Ward, and Wei-Jing Zhu. 2002.
\newblock \href {https://doi.org/10.3115/1073083.1073135} {{B}leu: a method for
  automatic evaluation of machine translation}.
\newblock In \emph{Proceedings of the 40th Annual Meeting of the Association
  for Computational Linguistics}, pages 311--318, Philadelphia, Pennsylvania,
  USA. Association for Computational Linguistics.

\bibitem[{Parikh et~al.(2020)Parikh, Wang, Gehrmann, Faruqui, Dhingra, Yang,
  and Das}]{parikh-etal-2020-totto}
Ankur Parikh, Xuezhi Wang, Sebastian Gehrmann, Manaal Faruqui, Bhuwan Dhingra,
  Diyi Yang, and Dipanjan Das. 2020.
\newblock \href {https://doi.org/10.18653/v1/2020.emnlp-main.89} {{ToTTo}: A
  controlled table-to-text generation dataset}.
\newblock In \emph{Proceedings of the 2020 Conference on Empirical Methods in
  Natural Language Processing (EMNLP)}, pages 1173--1186, Online. Association
  for Computational Linguistics.

\bibitem[{Perez-Beltrachini and
  Lapata(2021)}]{perez-beltrachini-lapata-2021-models}
Laura Perez-Beltrachini and Mirella Lapata. 2021.
\newblock \href {https://doi.org/10.18653/v1/2021.emnlp-main.742} {Models and
  datasets for cross-lingual summarisation}.
\newblock In \emph{Proceedings of the 2021 Conference on Empirical Methods in
  Natural Language Processing}, pages 9408--9423, Online and Punta Cana,
  Dominican Republic. Association for Computational Linguistics.

\bibitem[{Perez-Beltrachini et~al.(2019)Perez-Beltrachini, Liu, and
  Lapata}]{perez-beltrachini-etal-2019-generating}
Laura Perez-Beltrachini, Yang Liu, and Mirella Lapata. 2019.
\newblock \href {https://doi.org/10.18653/v1/P19-1504} {Generating summaries
  with topic templates and structured convolutional decoders}.
\newblock In \emph{Proceedings of the 57th Annual Meeting of the Association
  for Computational Linguistics}, pages 5107--5116, Florence, Italy.
  Association for Computational Linguistics.

\bibitem[{Puduppully et~al.(2019{\natexlab{a}})Puduppully, Dong, and
  Lapata}]{puduppully-etal-2019-data}
Ratish Puduppully, Li~Dong, and Mirella Lapata. 2019{\natexlab{a}}.
\newblock \href {https://doi.org/10.18653/v1/P19-1195} {Data-to-text generation
  with entity modeling}.
\newblock In \emph{Proceedings of the 57th Annual Meeting of the Association
  for Computational Linguistics}, pages 2023--2035, Florence, Italy.
  Association for Computational Linguistics.

\bibitem[{Puduppully and Lapata(2021)}]{puduppully-lapata-2021-data}
Ratish Puduppully and Mirella Lapata. 2021.
\newblock \href {https://doi.org/10.1162/tacl_a_00381} {Data-to-text generation
  with macro planning}.
\newblock \emph{Transactions of the Association for Computational Linguistics},
  9:510--527.

\bibitem[{Puduppully et~al.(2019{\natexlab{b}})Puduppully, Mallinson, and
  Lapata}]{puduppully-etal-2019-university}
Ratish Puduppully, Jonathan Mallinson, and Mirella Lapata. 2019{\natexlab{b}}.
\newblock \href {https://doi.org/10.18653/v1/D19-5630} {{U}niversity of
  {E}dinburgh{'}s submission to the document-level generation and translation
  shared task}.
\newblock In \emph{Proceedings of the 3rd Workshop on Neural Generation and
  Translation}, pages 268--272, Hong Kong. Association for Computational
  Linguistics.

\bibitem[{Pushkarna et~al.(2022)Pushkarna, Zaldivar, and
  Kjartansson}]{pushkarna2021datacardsplaybookupdated}
Mahima Pushkarna, Andrew Zaldivar, and Oddur Kjartansson. 2022.
\newblock \href {https://doi.org/10.48550/ARXIV.2204.01075} {Data cards:
  Purposeful and transparent dataset documentation for responsible ai}.

\bibitem[{Quan et~al.(2020)Quan, Zhang, Cao, Li, and
  Xiong}]{quan-etal-2020-risawoz}
Jun Quan, Shian Zhang, Qian Cao, Zizhong Li, and Deyi Xiong. 2020.
\newblock \href {https://doi.org/10.18653/v1/2020.emnlp-main.67} {{R}i{SAWOZ}:
  A large-scale multi-domain {W}izard-of-{O}z dataset with rich semantic
  annotations for task-oriented dialogue modeling}.
\newblock In \emph{Proceedings of the 2020 Conference on Empirical Methods in
  Natural Language Processing (EMNLP)}, pages 930--940, Online. Association for
  Computational Linguistics.

\bibitem[{Raji et~al.(2021)Raji, Denton, Bender, Hanna, and
  Paullada}]{DBLP:conf/nips/RajiDBHP21}
Inioluwa~Deborah Raji, Emily Denton, Emily~M. Bender, Alex Hanna, and
  Amandalynne Paullada. 2021.
\newblock \href
  {https://datasets-benchmarks-proceedings.neurips.cc/paper/2021/hash/084b6fbb10729ed4da8c3d3f5a3ae7c9-Abstract-round2.html}
  {{AI} and the everything in the whole wide world benchmark}.
\newblock In \emph{Proceedings of the Neural Information Processing Systems
  Track on Datasets and Benchmarks 1, NeurIPS Datasets and Benchmarks 2021,
  December 2021, virtual}.

\bibitem[{{Rajpurkar} et~al.(2016){Rajpurkar}, {Zhang}, {Lopyrev}, and
  {Liang}}]{2016arXiv160605250R}
Pranav {Rajpurkar}, Jian {Zhang}, Konstantin {Lopyrev}, and Percy {Liang}.
  2016.
\newblock \href {http://arxiv.org/abs/1606.05250} {{SQuAD: 100,000+ Questions
  for Machine Comprehension of Text}}.
\newblock \emph{arXiv e-prints}, page arXiv:1606.05250.

\bibitem[{Rastogi et~al.(2020)Rastogi, Zang, Sunkara, Gupta, and
  Khaitan}]{rastogi2020towards}
Abhinav Rastogi, Xiaoxue Zang, Srinivas Sunkara, Raghav Gupta, and Pranav
  Khaitan. 2020.
\newblock Towards scalable multi-domain conversational agents: The
  schema-guided dialogue dataset.
\newblock In \emph{Proceedings of the AAAI Conference on Artificial
  Intelligence}, volume~34, pages 8689--8696.

\bibitem[{Saleh et~al.(2019)Saleh, Berard, Calapodescu, and
  Besacier}]{saleh-etal-2019-naver}
Fahimeh Saleh, Alexandre Berard, Ioan Calapodescu, and Laurent Besacier. 2019.
\newblock \href {https://doi.org/10.18653/v1/D19-5631} {Naver labs {E}urope{'}s
  systems for the document-level generation and translation task at {WNGT}
  2019}.
\newblock In \emph{Proceedings of the 3rd Workshop on Neural Generation and
  Translation}, pages 273--279, Hong Kong. Association for Computational
  Linguistics.

\bibitem[{Scialom et~al.(2020)Scialom, Dray, Lamprier, Piwowarski, and
  Staiano}]{scialom2020mlsum}
Thomas Scialom, Paul-Alexis Dray, Sylvain Lamprier, Benjamin Piwowarski, and
  Jacopo Staiano. 2020.
\newblock Mlsum: The multilingual summarization corpus.
\newblock \emph{arXiv preprint arXiv:2004.14900}.

\bibitem[{Sellam et~al.(2020)Sellam, Das, and Parikh}]{sellam-etal-2020-bleurt}
Thibault Sellam, Dipanjan Das, and Ankur Parikh. 2020.
\newblock \href {https://doi.org/10.18653/v1/2020.acl-main.704} {{BLEURT}:
  Learning robust metrics for text generation}.
\newblock In \emph{Proceedings of the 58th Annual Meeting of the Association
  for Computational Linguistics}, pages 7881--7892, Online. Association for
  Computational Linguistics.

\bibitem[{Sulem et~al.(2018)Sulem, Abend, and Rappoport}]{sulem-etal-2018-bleu}
Elior Sulem, Omri Abend, and Ari Rappoport. 2018.
\newblock \href {https://doi.org/10.18653/v1/D18-1081} {{BLEU} is not suitable
  for the evaluation of text simplification}.
\newblock In \emph{Proceedings of the 2018 Conference on Empirical Methods in
  Natural Language Processing}, pages 738--744, Brussels, Belgium. Association
  for Computational Linguistics.

\bibitem[{Sun et~al.(2021)Sun, Hou, Wang, Zhang, and Wang}]{sun-etal-2021-d2s}
Edward Sun, Yufang Hou, Dakuo Wang, Yunfeng Zhang, and Nancy X.~R. Wang. 2021.
\newblock \href {https://doi.org/10.18653/v1/2021.naacl-main.111} {{D}2{S}:
  Document-to-slide generation via query-based text summarization}.
\newblock In \emph{Proceedings of the 2021 Conference of the North American
  Chapter of the Association for Computational Linguistics: Human Language
  Technologies}, pages 1405--1418, Online. Association for Computational
  Linguistics.

\bibitem[{Thomson et~al.(2020)Thomson, Reiter, and
  Sripada}]{thomson-etal-2020-sportsett}
Craig Thomson, Ehud Reiter, and Somayajulu Sripada. 2020.
\newblock \href {https://aclanthology.org/2020.intellang-1.4}
  {{S}port{S}ett:basketball - a robust and maintainable data-set for natural
  language generation}.
\newblock In \emph{Proceedings of the Workshop on Intelligent Information
  Processing and Natural Language Generation}, pages 32--40, Santiago de
  Compostela, Spain. Association for Computational Lingustics.

\bibitem[{Tiedemann and Nygaard(2004)}]{tiedemann-nygaard-2004-opus}
J{\"o}rg Tiedemann and Lars Nygaard. 2004.
\newblock \href {http://www.lrec-conf.org/proceedings/lrec2004/pdf/320.pdf}
  {The {OPUS} corpus - parallel and free: \url{http://logos.uio.no/opus}}.
\newblock In \emph{Proceedings of the Fourth International Conference on
  Language Resources and Evaluation ({LREC}{'}04)}, Lisbon, Portugal. European
  Language Resources Association (ELRA).

\bibitem[{Tonelli et~al.(2016)Tonelli, Aprosio, and
  Saltori}]{Tonelli2016SIMPITIKIAS}
Sara Tonelli, Alessio~Palmero Aprosio, and Francesca Saltori. 2016.
\newblock Simpitiki: a simplification corpus for italian.
\newblock In \emph{CLiC-it/EVALITA}.

\bibitem[{van~der Lee et~al.(2019)van~der Lee, Gatt, van Miltenburg, Wubben,
  and Krahmer}]{van-der-lee-etal-2019-best}
Chris van~der Lee, Albert Gatt, Emiel van Miltenburg, Sander Wubben, and Emiel
  Krahmer. 2019.
\newblock \href {https://doi.org/10.18653/v1/W19-8643} {Best practices for the
  human evaluation of automatically generated text}.
\newblock In \emph{Proceedings of the 12th International Conference on Natural
  Language Generation}, pages 355--368, Tokyo, Japan. Association for
  Computational Linguistics.

\bibitem[{Wang et~al.(2022)Wang, Pang, Chen, Phang, and
  Bowman}]{wang2022squality}
Alex Wang, Richard~Yuanzhe Pang, Angelica Chen, Jason Phang, and Samuel~R.
  Bowman. 2022.
\newblock S{Q}u{ALITY}: Building a long-document summarization dataset the hard
  way.
\newblock \emph{arXiv preprint 2205.11465}.

\bibitem[{Wang et~al.(2019)Wang, Pruksachatkun, Nangia, Singh, Michael, Hill,
  Levy, and Bowman}]{DBLP:conf/nips/WangPNSMHLB19}
Alex Wang, Yada Pruksachatkun, Nikita Nangia, Amanpreet Singh, Julian Michael,
  Felix Hill, Omer Levy, and Samuel~R. Bowman. 2019.
\newblock \href
  {https://proceedings.neurips.cc/paper/2019/hash/4496bf24afe7fab6f046bf4923da8de6-Abstract.html}
  {Superglue: {A} stickier benchmark for general-purpose language understanding
  systems}.
\newblock In \emph{Advances in Neural Information Processing Systems 32: Annual
  Conference on Neural Information Processing Systems 2019, NeurIPS 2019,
  December 8-14, 2019, Vancouver, BC, Canada}, pages 3261--3275.

\bibitem[{Wiseman et~al.(2017)Wiseman, Shieber, and
  Rush}]{wiseman-etal-2017-challenges}
Sam Wiseman, Stuart Shieber, and Alexander Rush. 2017.
\newblock \href {https://doi.org/10.18653/v1/D17-1239} {Challenges in
  data-to-document generation}.
\newblock In \emph{Proceedings of the 2017 Conference on Empirical Methods in
  Natural Language Processing}, pages 2253--2263, Copenhagen, Denmark.
  Association for Computational Linguistics.

\bibitem[{Xu et~al.(2016)Xu, Napoles, Pavlick, Chen, and
  Callison-Burch}]{xu-etal-2016-optimizing}
Wei Xu, Courtney Napoles, Ellie Pavlick, Quanze Chen, and Chris Callison-Burch.
  2016.
\newblock \href {https://doi.org/10.1162/tacl_a_00107} {Optimizing statistical
  machine translation for text simplification}.
\newblock \emph{Transactions of the Association for Computational Linguistics},
  4:401--415.

\bibitem[{Xu et~al.(2022)Xu, Wang, Yu, Ritchie, Yao, Wu, Zhang, Li, Bradford,
  Sun, Hoang, Sang, Hou, Ma, Yang, Peng, Yu, and
  Warschauer}]{xu2022fairytaleqa}
Ying Xu, Dakuo Wang, Mo~Yu, Daniel Ritchie, Bingsheng Yao, Tongshuang Wu, Zheng
  Zhang, Toby Jia-Jun Li, Nora Bradford, Branda Sun, Tran~Bao Hoang, Yisi Sang,
  Yufang Hou, Xiaojuan Ma, Diyi Yang, Nanyun Peng, Zhou Yu, and Mark
  Warschauer. 2022.
\newblock Fantastic questions and where to find them: Fairytale{QA} -- an
  authentic dataset for narrative comprehension.
\newblock Association for Computational Linguistics.

\bibitem[{Xue et~al.(2021)Xue, Constant, Roberts, Kale, Al-Rfou, Siddhant,
  Barua, and Raffel}]{xue-etal-2021-mt5}
Linting Xue, Noah Constant, Adam Roberts, Mihir Kale, Rami Al-Rfou, Aditya
  Siddhant, Aditya Barua, and Colin Raffel. 2021.
\newblock \href {https://doi.org/10.18653/v1/2021.naacl-main.41} {m{T}5: A
  massively multilingual pre-trained text-to-text transformer}.
\newblock In \emph{Proceedings of the 2021 Conference of the North American
  Chapter of the Association for Computational Linguistics: Human Language
  Technologies}, pages 483--498, Online. Association for Computational
  Linguistics.

\bibitem[{Zhang et~al.(2020)Zhang, Zhu, Brahma, and Li}]{zhang-etal-2020-small}
Li~Zhang, Huaiyu Zhu, Siddhartha Brahma, and Yunyao Li. 2020.
\newblock \href {https://doi.org/10.18653/v1/2020.emnlp-main.91} {Small but
  mighty: New benchmarks for split and rephrase}.
\newblock In \emph{Proceedings of the 2020 Conference on Empirical Methods in
  Natural Language Processing (EMNLP)}, pages 1198--1205, Online. Association
  for Computational Linguistics.

\bibitem[{Zhu et~al.(2020)Zhu, Huang, Zhang, Zhu, and
  Huang}]{zhu-etal-2020-crosswoz}
Qi~Zhu, Kaili Huang, Zheng Zhang, Xiaoyan Zhu, and Minlie Huang. 2020.
\newblock \href {https://doi.org/10.1162/tacl_a_00314} {{C}ross{WOZ}: A
  large-scale {C}hinese cross-domain task-oriented dialogue dataset}.
\newblock \emph{Transactions of the Association for Computational Linguistics},
  8:281--295.

\end{thebibliography}

\begin{table*}[ht!]
\tiny
\begin{tabular}{@{}p{2.5cm}p{2.6cm}llrrrr@{}}
\toprule
Dataset & Citation & Task & Language(s) & Taxonomy & Size & Input Length & Output Length \\ \midrule
ART & \cite{Bhagavatula2020Abductive} & Reasoning & en & 5 & 50k & 138 & 41  \\
BiSECT & \cite{kim-etal-2021-bisect} & Simplification & en, de, es, fr & 5 & 200k--1M & 266--434 & 224--387 \\
Cochrane & \cite{devaraj-etal-2021-paragraph} & Simplification & en & 5 & 3.5k &  &  \\
CommonGen & \cite{lin-etal-2020-commongen} & Data-to-Text & en & 5 & 70k & 80 &  \\
Conversational Weather & \cite{balakrishnan-etal-2019-constrained} & Response Generation & en & 5 & 25k & 417 & 315 \\
CrossWOZ & \cite{zhu-etal-2020-crosswoz} & Response Generation & zh & 5 & 5k &  &  \\
CS Restaurants & \cite{dusek-jurcicek-2019-neural} & Response Generation & cs & 4 & 3.5k & 70 & 58  \\
DART & \cite{nan-etal-2021-dart} & Data-to-Text & en  & 5 & 60k &  &  \\
DSTC 10 & \cite{Kim2021HowRR} & Data-to-Text & en & 5 & 20k & 1337 & 95 \\
E2E NLG & \cite{novikova-etal-2017-e2e,DUSEK2020123,dusek-etal-2019-semantic} & Data-to-Text & en & 5 & 35k & 146 & 135 \\
FairytaleQA & \cite{xu2022fairytaleqa} & Question Geneartion & en & 5 & 8.5k & 335 & 15.9 \\
IndoNLG & \cite{cahyawijaya-etal-2021-indonlg} & Summarization & id, jv, su & 1--3 & 14k--200k & 2021 & 456 \\
MLB &  \cite{puduppully-etal-2019-data} & Data-to-Text & en & 5 & 23k & 24665 & 2580 \\
MLSum & \cite{scialom2020mlsum} & Summarization & es, de & 5 & 220k--250k & 4152 & 147 \\
Opusparcus & \cite{creutz:lrec2018} & Paraphrasing & de, en, fi, fr, ru, sv & 4--5 & 0--35M & &  \\
OrangeSum &  \cite{kamal-eddine-etal-2021-barthez} & Summarization & fr & 5 & 21k--30k & 1984 & 138 \\
RiSAWOZ & \cite{quan-etal-2020-risawoz} & Response Generation & zh & 5 & 10k &  &  \\
RotoWire En-De & \cite{wiseman-etal-2017-challenges, hayashi-etal-2019-findings} & Data-to-Text & en, de & 5 & 242 &  &  \\
Schema-Guided Dialog & \cite{rastogi2020towards} & Response Generation & en & 5 & 165k & 188 & 51 \\
SciDuet & \cite{sun-etal-2021-d2s} & Slide Generation & en & 5 & 2k &  &  \\
SIMPITIKI & \cite{Tonelli2016SIMPITIKIAS} & Simplification & it & 4 & 815 &  &  \\
SportSett & \cite{thomson-etal-2020-sportsett} & Data-to-Text & en & 5 & 3.7k & 5990 & 1620 \\
Squad V2 & \cite{2016arXiv160605250R} & Question Generation & en & 5 & 120k & 768 & 55 \\
SQuALITY v1.1 & \cite{wang2022squality} & Summarization & en & & 2500 & 5000 & 227 \\
Surface Realization ST 2020 & \cite{mille-etal-2020-third} & Data-to-Text & ar, en, es, fr, hi, in & 3--5 & 250k & 892 & 126 \\
& & & ko, ja, pt, ru, zh & & & &\\
TaskMaster &  \cite{byrne-etal-2019-taskmaster} & Response Generation & en & 5 & 190k & 972 & 55 \\
ToTTo & \cite{parikh-etal-2020-totto} & Data-to-Text & en & 5 & 120k & 357 &  \\
Turku Hockey & \cite{kanerva-etal-2019-template} & Data-to-Text & fi & 4 & 2.7k--6.1k & 158 & 58 \\
Turku Paraphrase & \cite{kanerva-etal-2021-finnish} & Paraphrasing & fi & 4 & 81k--170k & 87 & 47 \\
ViGGo & \cite{juraska-etal-2019-viggo} & Data-to-Text & en & 5 & 5.1k & 120 & 109 \\
WebNLG & \cite{gardent-etal-2017-creating,gardent-etal-2017-webnlg}  & Data-to-Text & en, ru & 4--5 & 14k--35k & 169.5 & 157 \\
WikiAuto \\ +ASSET/TURK/Split\&Rephrase & \cite{jiang-etal-2020-neural,alva-manchego-etal-2020-asset,xu-etal-2016-optimizing,zhang-etal-2020-small} & Simplification & en & 5 & 480k &  &  \\
WikiCatSum & \cite{perez-beltrachini-etal-2019-generating} & Summarization & en & 5 & 48k & 43527 & 256 \\
WikiLingua & \cite{ladhak-etal-2020-wikilingua} & Summarization & ar, cs, de, en, es, fr, & 3--5 & 5k--3.8M & 1607--4650 & 159--489\\
& & & hi, id, it, ja, ko, nl, & & & & \\
& & & pt, ru, th, tr, vi, zh &  &  & 2244.5 & 200.5 \\
XLSum & \cite{hasan-etal-2021-xl} & Summarization & om, fr, am, ar, az, bn, & 0--5 & 1.3k--300k & 1470--9924 & 137--614 \\
& & &  cy, en, es, gd, fa, & & & & \\
& & &  gu, ha, hi, ig, id, ja, & & & & \\
& & &  ko, ky, mr, my, ne,  & & & & \\
& & & ps,  pcm, pt, pa, rn, ru,  & & & & \\
& & & sr, si, so, sw, ta, te, & & & & \\
& & &  th, ti, tr, uk, ur, uz, & & & & \\
& & & vi, yo, zh-CN, zh-TW &  &  & 3486.5 & 237 \\
XSum & \cite{narayan-etal-2018-dont} & Summarization & en & 5 & 23k & 1845 & 153  \\
XWikis & \cite{perez-beltrachini-lapata-2021-models} & Summarization & en, de, fr, cs & 4-5 & 44k--461k & 1743 & 102  \\
\bottomrule
\end{tabular}
\caption{Detailed information about all the datasets currently supported in GEM. We present the name of the dataset, the paper(s) in which the dataset was introduced, the NLG task it performs, the languages the dataset caters to and their resourcedness taxonomy class, the size of the training set (rounded), and the lengths of input and output.}
\label{tab:datasets}
\end{table*}

\newpage

\appendix

\section{Dataset Overviews}
\label{app:overview}
We provide a detailed overview of all the supported datasets in Table \ref{tab:datasets}. Input and output lengths are reported in number of tokens according to the mT5 tokenizer~\citep{xue-etal-2021-mt5}. When multiple configurations for a dataset are available, we report the median of the sizes and lengths.










\section{Changes to Datasets}

\subsection{BiSECT}


The original released \emph{BiSECT} \cite{kim-etal-2021-bisect} training, validation, and test splits are maintained to ensure a fair comparison. Note that the original BiSECT test set was created by manually selecting 583 high-quality Split and Rephrase instances from 1000 random source-target pairs sampled from the EMEA and JRC-Acquis corpora from the OPUS parallel corpus \cite{tiedemann-nygaard-2004-opus}.

As the first challenge set, we include the \emph{HSPLIT-Wiki} test set, containing 359 pairs \cite{sulem-etal-2018-bleu}. For each complex sentence, there are four reference splits; To ensure replicability, as reference splits, we again follow the original \emph{BiSECT} paper and present only the references from \emph{HSplit2-full}. In addition to the two evaluation sets used in the original BiSECT paper, we also introduce a second challenge set. For this, we initially consider all 7,293 pairs from the EMEA and JRC-Acquis corpora. From there, we classify each pair using the classification algorithm from Section 4.2 of the original BiSECT paper. The three classes are as follows:

\begin{enumerate}
\item \textbf{Direct Insertion}: when a long sentence \textit{l} contains two independent clauses and requires only minor changes in order to make a fluent and meaning-preserving split \textit{s}.
\item \textbf{Changes near Split}, when \textit{l} contains one independent and one dependent clause, but modifications are restricted to the region where \textit{l} is split.
\item \textbf{Changes across Sentences}, where major changes are required throughout \textit{l} in order to create a fluent split \textit{s}.
\end{enumerate}

We keep only pairs labeled as Type 3, and after filtering out pairs with significant length differences (signaling potential content addition/deletion), we present a second challenge set of 1,798 pairs.

\subsection{FairytaleQA}
The original release of FairytaleQA \cite{xu2022fairytaleqa} used separate files to store the fairytale story content and experts-labeled QA-pairs. It provided baseline benchmarks on both Question Answering and Question Generation tasks. In GEMv2, we re-organize the data to be specifically prepared for the Question Generation task. The original dataset contains 2 answers created by different annotators in the evaluation and test splits, but we only take the first answer into consideration for the Question Generation task. The input for this task would be the concatenation of each answer labeled by human experts and the related story section(s), and the output target would be the corresponding question labeled by human experts.

\subsection{MLB Data to Text}
We follow the serialization format introduced in \cite{puduppully-lapata-2021-data} for the linearized\_input field. Specifically, we serialize the home team records, the visiting team records, and the player records. We next serialize the records of the innings in chronological order.

\subsection{Opusparcus}


Compared to the original release of Opusparcus \cite{creutz:lrec2018}, available through the Language Bank of Finland,\footnote{\url{https://www.kielipankki.fi/corpora/opusparcus/}} the GEMv2 release contains a few additions to facilitate the use of this resource:

The validation and test sets now come in two versions, the so-called \emph{regular} validation and test sets and the \emph{full} sets. The regular sets only contain sentence pairs that qualify as paraphrases. The full sets are the original sets from the original release, which contain all sentence pairs successfully annotated by the annotators, including the sentence pairs that were rejected as paraphrases. The validation sets were called development sets in the original release.

The training sets are orders of magnitudes larger than the validation and test sets. Therefore the training sets have not been annotated manually and the true paraphrase status of each entry is unknown. In the original release, each training set entry is accompanied by an automatically calculated ranking score, which reflects how likely that entry contains a true paraphrase pair. The entries are ordered in the data, best first, worst last. If you use the original release, you need to control yourself how large and how clean a portion of the training data you will use.

In the GEMv2 release, the training sets come in predefined subsets. Using the so-called \emph{quality} parameter, the user can control for the estimated proportion (in percent) of true paraphrases in the retrieved training subset. Allowed quality values range between 60 and 100, in increments of 5 (60, 65, 70, ..., 100). A value of 60 means that 60\,\% of the sentence pairs in the training set are estimated to be true paraphrases (and the remaining 40\,\% are not). A higher value produces a smaller but cleaner set. The smaller sets are subsets of the larger sets, such that the quality=95 set is a subset of quality=90, which is a subset of quality=85, and so on. Depending on this parameter, the dataset can fall into all resourcedness categories in Figure~\ref{fig:overview}.

\subsection{\textsc{RotoWire}\_English-German}
We introduce a field linearized\_input, which serializes the input table into a string. We follow a serialization format similar to that of \citet{saleh-etal-2019-naver}. More specifically, we serialize all the records of the home team followed by that of the visiting team. We next serialize the records of the players of the home team followed by that of the visiting team. We rank the players by points in descending order. In addition, we add information about the relative rank of a player within a team following \citet{puduppully-etal-2019-university}.

\subsection{SciDuet}
The original released \emph{SciDuet} \cite{sun-etal-2021-d2s} uses two json files to store paper information and slide information, respectively. In GEMv2, we merge these two files and reorganize the structure so that each data instance contains the complete input (i.e., paper title/abstract/section headers/section content, as well as slide title) and output (i.e., slide text content). In addition, we introduce a new challenging dataset in GEMv2 by removing slides if their titles match with any section headers from the corresponding paper.

\subsection{SIMPITIKI}
The original release of SIMPITIKI \cite{Tonelli2016SIMPITIKIAS} includes two xml files, corresponding to the version 1 and version 2 respectively. The second version has better sentence boundaries. However, no training, validation and test splits were officially proposed for both release. In GEM, we randomly and independently split both xml files into training, validation and test sets. Note that version 1 and version 2 have different splits. We also generated challenge sets were some simplification transformations in the test set are not part of the training set and thus unseen in the training phase. Then, as SIMPITIKI leverages data from Wikipedia and the Municipality of Trento corpora, we further propose splits based on the respective data source.

\subsection{SportSett Basketball}
Similar to MLB Data-to-Text, SportSett also follows the serialization format introduced in \cite{puduppully-lapata-2021-data} for the linearized\_input field. The serialisation starts with current game's information such as date and venue of the game. This is followed with both team's information (line-scores) including their next game's information as well. Finally, the players' information (box-scores) is serialised, starting with home team's players and then visiting team's players.

\subsection{squad\_v2}
SQuAD2.0 \cite{2016arXiv160605250R} combines the 100,000 questions in SQuAD1.1 with over 50,000 unanswerable questions written adversarially by crowdworkers to look similar to answerable ones. The original SQuAD2.0 dataset has only training and dev (validation) splits. A new test split is created from the train split and added as part of the squad\_v2 dataset.

\subsection{Taskmaster-3}
According to \citet{byrne-etal-2021-tickettalk}, the Taskmaster-3 (also called TicketTalk) dataset consists of 23,789 movie ticketing dialogs, where the customer's goal is to purchase tickets after deciding on theater, time, movie name, number of tickets, and date, or opt out of the transaction.
This collection was created using the "self-dialog" method, i.e., a single, crowd-sourced worker is paid to create a conversation writing turns for both speakers- the customer and the ticketing agent.

\subsection{Turku Hockey}

To ease the use of the data, in addition to the game-level structuring as used in the original Turku Hockey data release~\cite{kanerva-etal-2019-template}, we provide a simplified event-level structuring. In the event-level generation, the structured input data is linearized to string representation separately for each game event, and  the task objective is thus to generate the description separately for each game event directly using the linearized input representation. In comparison, the objective of the game-level generation is to process the structured data for the entire game at once, and generate descriptions for all relevant events. The linearized event inputs are produced using similar approach as described in the original paper.

\subsection{Turku Paraphrase}

In GEMv2, the Turku Paraphrase data can be loaded with three different configurations, \emph{plain}, \emph{classification}, and \emph{generation}. While the \emph{plain} configuration models the data similarly to the original release, the two other options directly applies several transformations beneficial for the named task. In \emph{classification} each example is provided using both \emph{(text1, text2, label)} and \emph{(text2, text1, label)} ordering, as paraphrase classification does not depend on the order of the given statements. In cases with a directionality annotation in the paraphrase pair, the label is flipped accordingly when creating the additional examples. In generation, on the other hand, the data is pre-processed to include only examples suitable for the paraphrase generation task, therefore discarding, e.g.,  negative and highly context dependent examples, which does not fit the generation task as such. In addition, the examples with annotated directionality (one statements being more detailed than the other, for instance one mentioning \emph{a woman} while the other \emph{a person}), the example is always provided using ordering where the input is more detailed and the output more general in order to prevent model hallucination (model learning to generate facts not present in the input). For more details about the annotated labels and the directionality, see \citet{kanerva-etal-2020-turku}.

\subsection{WikiLingua}
The original release of WikiLingua \cite{ladhak-etal-2020-wikilingua} released a dataset of article-summary pairs in $18$ languages, but had only created train/val/test splits for $4$ langauge pairs (es-en, tr-en, ru-en, vi-en), for the purposes of crosslingual evaluation. As part of GEMv1, we created train/val/test splits for all $18$ languages. To further facilitate building multilingual and crosslingual models for all $18$ languages, the GEMv2 release contains the following changes to the GEMv1 release:

In the original WikiLingua release, each document-summary pair in any of the $17$ non-English languages has a corresponding parallel document-summary pair in English. A given English document-summary pair can have parallel document-summary pairs in multiple languages. In order to facilitate crosslingual experiments across all language pairs, for the GEMv2 release, we align document-summary pairs across the other $17$ languages via English. For example, if a given document-summary pair in English has corresponding parallel pairs in Turkish and Vietnamese, we can then align these to get Turkish-Vietnamese parallel pairs. As a result, in addition to supporting all the functionality in GEMv1, the v2 loader allows the user to specify and load crosslingual data for any language pair in the dataset.

In addition to the original evaluation sets (val and test), we also have sub-sampled versions in order to facilitate faster development cycles. To create the sub-sampled versions, for each evaluation set, we randomly sample $3,000$ instances.\footnote{Evaluation sets that have fewer than $3,000$ instances were not sub-sampled.}

We further clean the dataset by removing payloads for thumbnails that were scraped into the document and summary texts and we filter out all instances with a summary length longer than 60\% of the input document length. This removes around 5\% of the data.

\section{Limitations}

As discussed in the main part of the paper, GEMv2 aims to avoid any explicit curation decisions about inclusion and exclusion of datasets beyond licensing and consent. This is a change from the originally set out strict inclusion criteria based on dataset quality. The reason for this is that the entire research community should be the authority to decide whether a dataset is useful and what it is useful for. For example, a dataset with noisy outputs may still be useful to study hallucination avoidance methods.
However, this change has implications on how dataset deprecation needs to be handled, in particular for datasets with newly found issues or datasets with better alternatives.
Documenting issues and alternatives using the data cards is thus becoming more important in GEMv2 and we encourage researchers to update data cards.
Another side effect of positioning GEMv2 as infrastructure that support dataset creators is a decreased risk of erasure. All our documentation and dataset loaders center the work of the creators to encourage users to cite the datasets they use.

Another open issue that we have been working on is the interplay between multilingualism and metrics. We now support multiple languages for which no NLG metrics have been tested, and for which our tokenization schemes may be inappropriate. The freedom to combine every dataset with every metric may lead to flawed evaluations. In addition, some datasets were released with specific metrics that we do not support yet.

A final issue we want to point out is the lack of discussion of human evaluation in this overview paper which we omitted for brevity. Human evaluation does not scale and every task requires its own evaluation approach. We have thus taken the approach to develop better human evaluation for only a subset of tasks, solving issues pointed out by  \cite{howcroft-etal-2020-twenty} and \citet{van-der-lee-etal-2019-best}, and we will release detailed instructions separately.

\section{Contribution Statements}

Organizing GEM would not be possible without community contributions and the mutual goal of improving NLG and its evaluation. To give proper credit to all contributors, this section lists the involvements of all co-authors. Besides the detailed list, everyone contributed to discussion sessions, made dataset suggestions, and participated in proof reading the final paper.

\paragraph{Dataset Loaders}

The new data loaders and associated data cards were created by the following people:

\noindent \textit{ART}: Chandra Bhagavatula, Nico Daheim, Aman Madaan \\
\noindent \textit{BiSect}: Jenny Chim, Reno Kriz \\
\noindent \textit{Conversational Weather}: Vipul Raheja, Michael White \\
\noindent \textit{CrossWOZ}: Qi Zhu \\
\noindent \textit{DSTC10}: Nico Daheim, Di Jin, Alexandros Papangelis  \\
\noindent \textit{FairyTaleQA}: Bingsheng Yao \\
\noindent \textit{IndoNLG}: Bryan Wilie, Samuel Cahyawijaya, Genta Indra Winata \\
\noindent \textit{MLB}: Ratish Puduppully \\
\noindent \textit{Opusparcus}: Mathias Creutz \\
\noindent \textit{OrangeSum}: Moussa Kamal Eddine \\
\noindent \textit{RiSAWOZ}: Tianhao Shen, Deyi Xiong, Chaobin You \\
\noindent \textit{RotoWire En-De}: Hiroaki Hayashi, Ratish Puduppully \\
\noindent \textit{SciDuet}: Yufang Hou, Dakuo Wang \\
\noindent \textit{SIMPITIKI}: Sebastien Montella, Vipul Raheja \\
\noindent \textit{Split and Rephrase}: Cristina Garbacea, Reno Kriz, Li Zhang \\
\noindent \textit{SportSett}: Craig Thomson, Ashish Upadhyay \\
\noindent \textit{Squad V2}: Abinaya Mahendiran \\
\noindent \textit{SQuALITY}: Alex Wang \\
\noindent \textit{Surface Realisation ST}: Bernd Bohnet, Simon Mille \\
\noindent \textit{TaskMaster}: Tosin Adewumi \\
\noindent \textit{ToTTo (port)}: Abinaya Mahendiran \\
\noindent \textit{Turku Hockey}: Filip Ginter, Jenna Kanerva \\
\noindent \textit{Turku Paraphrase}: Filip Ginter, Jenna Kanerva \\
\noindent \textit{ViGGo}: Juraj Juraska, Aman Madaan \\
\noindent \textit{WikiCatSum}: Ronald Cardenas Acosta, Laura Perez-Beltrachini \\
\noindent \textit{WikiLingua (port)}: Jenny Chim, Faisal Ladhak \\
\noindent \textit{XLSum}: Abhik Bhattacharjee, Tahmid Hasan, Rifat Shahriyar \\
\noindent \textit{XSum (port)}: Abinaya Mahendiran \\
\noindent \textit{XWikis}: Ronald Cardenas Acosta, Laura Perez-Beltrachini \\

\noindent Lewis Tunstall designed and implemented the infrastructure to host GEMv2 on the Hugging Face Hub. Sebastian Gehrmann addressed the remaining loader issues and ported the remaining GEMv1 datasets. Anna Shvets developed dataset-agnostic bias detection filters.
Simon Mille coordinated progress during the hackathon.

\paragraph{Documentation}
The updated tutorials for using GEM and adding new data loaders were developed and tested by Jenny Chim, Paul Pu Liang, and Anna Shvets.

\paragraph{Data Cards}
The questions in the revised data card template were created during sessions led by Mahima Pushkarna with the help of Yacine Jernite, Angelina McMillan-Major, Nishant Subramani, Pawan Sasanka Ammanamanchi, and Sebastian Gehrmann. The collection tool was implemented by Yacine Jernite and Sebastian Gehrmann.
The data card rendering tool was developed by Vivian Tsai and Mahima Pushkarna.

\paragraph{Human Evaluation}
The human evaluation working group is led by João Sedoc. Its members include Jenny Chim, Elizabeth Clark, Daniel Deutsch, Kaustubh Dhole, Sebastian Gehrmann, Yufang Hou, Yixin Liu, Saad Mahamood, Simon Mille, Vitaly Nikolaev, Salomey Osei, Dragomir Radev, Yisi Sang, and Alex Wang.

\paragraph{Metrics}
The metrics library, originally developed for GEMv1, was extended by Jordan Clive, Nico Daheim, Daniel Deutsch, Ondrej Dusek, Aman Madaan, Joshua Maynez, Vikas Raunak, Leonardo F. R. Ribeiro, and Anna Shvets.

\paragraph{Paper Writing and Analyses}
Sebastian Gehrmann led the writing of the paper. Abinaya Mahendiran and Jekaterina Novikova contributed analyses that were used to create Figure~\ref{fig:overview} and Table~\ref{tab:datasets}.

\paragraph{Submission Infrastructure}
Lewis Tunstall led the development of the submission infrastructure. Hendrik Strobelt led the extension of the result visualization tool to ensure compatibility with the new submission system.

\paragraph{Baselines}
Additional baseline results were provided by Tosin Adewumi, Mihir Sanjay Kale, Joshua Maynez, and Leonardo F. R. Ribeiro.

\end{document}